\documentclass[runningheads]{llncs}

\usepackage{eccv}

\usepackage[accsupp]{axessibility}  %

\usepackage{eccvabbrv}

\def\meth{HOPformer\xspace}
\def\fitname{EC-fit\xspace}
\def\methfull{\underline{H}and-\underline{O}bject \underline{P}ose trans\underline{former}\xspace}
\def\dataset{EPIC-Contact\xspace}
\newcommand{\supmat}{Sup.~Mat.\xspace}

\definecolor{cvprblue}{rgb}{0.21,0.49,0.74}
\newcommand{\myfirstpara}[1]{\noindent \textbf{#1:}}

\newcommand{\mypara}[1]{\vspace{0.2em} \myfirstpara{#1}}
\usepackage{booktabs}
\usepackage[inline]{enumitem}
\usepackage{multirow}
\usepackage{adjustbox}
\usepackage[utf8]{inputenc}
\usepackage[most]{tcolorbox}
\usepackage[table]{xcolor}

\usepackage{graphicx}
\usepackage{subfiles}
\usepackage[most]{tcolorbox}

\newtcolorbox{promptbox}{
  breakable,
  enhanced,
  colback=gray!5,
  colframe=gray!40,
  arc=2mm,
  boxrule=0.4pt,
  left=6pt,
  right=6pt,
  top=6pt,
  bottom=6pt,
  fonttitle=\bfseries,
}

\definecolor{chgRed}{HTML}{B00020}
\definecolor{chgBlack}{HTML}{000000}
\newcommand{\chg}[1]{\textcolor{chgBlack}{#1}}

\usepackage[pagebackref,breaklinks,colorlinks,allcolors=cvprblue]{hyperref}

\usepackage{orcidlink}

\begin{document}

\title{Towards in-the-wild Egocentric 3D Hand-Object Pose Estimation}

\titlerunning{\dataset \& \meth}

\author{Siddhant Bansal\inst{1} \quad
Zhifan Zhu\inst{1} \quad
Shashank Tripathi\inst{2} \quad
Jiahe Zhao\inst{1} \quad \\
Michael J. Black\inst{2} \quad \quad
Dima Damen\inst{1}}

\authorrunning{S. Bansal et al.}

\institute{University of Bristol, Bristol, United Kingdom
\\
\and
Max Planck Institute for Intelligent Systems, Tübingen, Germany\\
\href{https://sid2697.github.io/epic-contact/}{https://sid2697.github.io/epic-contact}
}

\maketitle

\begin{abstract}
Estimating accurate 3D hand–object pose from in-the-wild egocentric RGB remains challenging due to severe occlusions and ambiguous contact. 
Existing learning-based methods often struggle to generalise to in-the-wild scenes and 
are limited by the scarcity of supervision.
We address these issues with two contributions. First, we introduce \dataset, an in-the-wild egocentric dataset of $2.3$K clips ($62.3$K frames) with dense, bijective 3D hand–object contact correspondences and posed meshes.
Second, we propose \meth, an end-to-end transformer that jointly predicts bi-manual hand and object pose in a single forward pass. 
A cross-attention decoder conditions object features on hand priors, producing robust pose estimation. 
We test \meth on the in-lab 3D dataset, ARCTIC, as well as our newly introduced \dataset dataset. \meth reaches $82.4\%$ success rate on ARCTIC (+$6.2$ pts over current SOTA).
On \dataset, it nearly doubles the success rate while reducing contact deviation by $75\%$. 
\dataset, \meth code and checkpoints are released: \href{https://sid2697.github.io/epic-contact}{https://sid2697.github.io/epic-contact}.
\end{abstract}

\section{Introduction}
\label{sec:intro}

\begin{figure}[t]
    \centering
    \includegraphics[width=\linewidth]{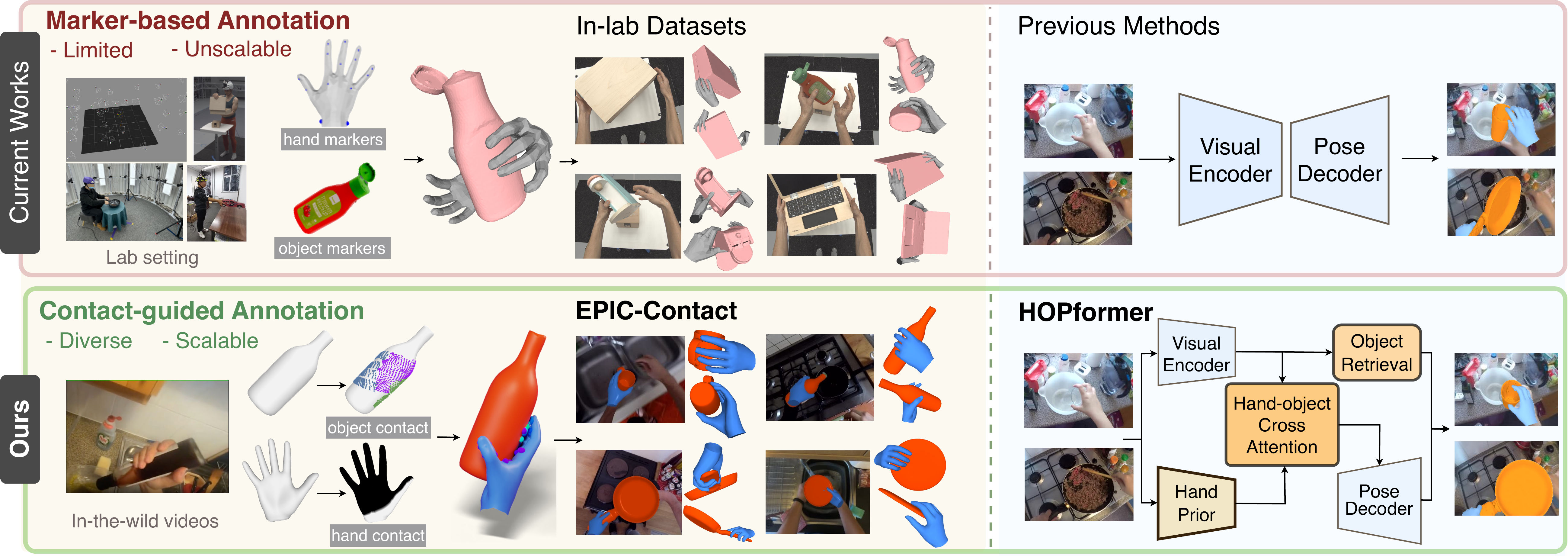}
    \caption{(\textit{Left}) We introduce \dataset, an in-the-wild egocentric dataset for 3D hand-object pose estimation.
    Unlike typical in-lab MoCap datasets that require specialised equipment and capture limited backgrounds/object instances, EPIC-Contact provides diverse, cluttered real-world interactions with posed 3D hand–object meshes derived from dense, bijective contact annotations.
    (\textit{Right}) Existing learning-based approaches do not leverage strong hand priors and hence do not generalise to in-the-wild scenarios.
    In contrast, our proposed \meth network enriches object features with hand priors to achieve superior performance on in-the-wild data.}
    \vspace{-18pt}
    \label{fig:teaser}
\end{figure}

We routinely use our two hands to interact with the physical world.
Everyday activities like washing dishes at the sink, grasping a bottle, or moving cookware on the stove require precise coordination between hands and objects, highlighting the remarkable dexterity and adaptability of the human hand.
Modelling such interactions is essential for human-centric applications such as AR/VR, robotics, human-computer interaction/collaboration, and assistive technologies.
Yet most existing work relies on controlled, scripted settings that fail to capture the complexity of real-world use~\cite{cao2021reconstructing, hasson2021towards, patel2022learning,fan2023arctic,AbouZeid2023JointTransformer}.
In this work, we aim to recover the pose of both the hands and the object in 3D from a single forward pass, including from images sourced from unscripted egocentric videos (\cref{fig:teaser}).

The task is inherently challenging due to the wide diversity of objects and hand poses commonly present in hand-object interaction scenarios.
The difficulty is further compounded by strong mutual occlusions and ambiguous contact regions.
Existing methods focus on controlled (in-lab) settings~\cite{fan2023arctic,banerjee2024hot3d,HOI4D,h2odataset,yang2022oakink,dexycb} 
and as shown in \cref{fig:teaser}~(\textit{Right}), often struggle in the wild~\cite{cao2021reconstructing, hasson2021towards, patel2022learning}. 

A central bottleneck is the availability of supervision.
Training (or evaluating) on in-the-wild interactions requires images paired with ground-truth 3D poses of both the hands and the object.
As highlighted in \cref{fig:teaser} (\textit{Left}),
such 3D ground-truth has, to date, been acquired using expensive motion capture (MoCap)~\cite{fan2023arctic, grab, yang2022oakink, HOI4D, banerjee2024hot3d, h2odataset}, which requires careful calibration, making it unsuitable for wider adoption.
Such data typically has limited,
minimalist uncluttered backgrounds that do not reflect the complexity of real-world egocentric scenes~\cite{fan2023arctic,banerjee2024hot3d,HOI4D,h2odataset}.

To move beyond these limitations, we propose a novel approach to 
obtain 3D annotations for hand-object interaction in images, thus enabling scalable supervision. 
Specifically, we introduce \textit{\dataset}, a dataset of $2.3$K egocentric video clips, capturing hand-object grasps with $9$ object categories, which we annotate with contact vertices and paired 3D hand-object meshes (\cref{fig:teaser}, \textit{Left}).
\dataset captures diverse backgrounds and cluttered environments, where objects may be small or transparent, heavily occluded, in challenging natural interactions.
Our key idea is to annotate bijective 3D contact on \emph{both} the hands and the object, in line with prior work on full-body contact~\cite{tripathi2023deco, cseke_tripathi_2025_pico}.
Once we have bijective contact points,
we use an optimisation pipeline to translate these annotations into posed 3D hand-object meshes.
With this approach, we obtain $62.3$K annotated frames, 
a one-of-a-kind dataset for training and evaluation.

While learning-based 3D hand pose estimation has progressed rapidly
~\cite{boukhayma20193d,pavlakos2024reconstructing_hamer, Potamias_2025_CVPR_wilor,rong2021frankmocap,prakash20243d,kulon2019single,kulon2020weakly}, performing well in diverse and challenging scenarios, the same cannot be said of 
joint hand-object pose estimation methods~\cite{fan2023arctic,AbouZeid2023JointTransformer}. 
Such methods fail to model the interactions between hand and object poses, impacting their performance as shown in \cref{fig:teaser} (\textit{Right}).

Our intuition is to leverage robust data-driven 3D hand reconstruction methods~\cite{pavlakos2024reconstructing_hamer, Potamias_2025_CVPR_wilor} to guide joint 3D hand and object pose estimation.
We propose \meth, an end-to-end learning framework for joint 3D bi-manual hand-object pose estimation from an RGB image in a single pass (\cref{fig:teaser}, \textit{Right}).
Specifically, we design a cross-attention framework that injects structural hand priors into object features, creating a powerful set of interaction features.
\meth\ generalises well across both in-lab~\cite{fan2023arctic} and in-the-wild images, 
and achieves state-of-the-art results.
Together, \dataset and \meth represent a step towards scalable, robust 3D hand-object pose estimation.
To summarise, our contributions are:
\begin{enumerate}[leftmargin=*,noitemsep, nolistsep]
    \item We annotate 3D hand-object contact points on $2.3$K video clips from the EPIC-Kitchens~\cite{Damen2022RESCALING,zhu2024grip} dataset and propose \dataset.
    This enables training and evaluation of hand-object pose estimation methods on challenging in-the-wild images. 
    \item We propose a learning-based transformer model, \methfull (\meth), that leverages pre-trained hand priors to regress the pose of the two hands and the object in a single forward pass. 
    \item \meth achieves state-of-the-art results on ARCTIC~\cite{fan2023arctic} and \dataset, outperforming prior work on most metrics by large margins.
\end{enumerate}

\section{Related Works}

\mypara{Hand-Object Pose Estimation}
Jointly estimating the 3D pose of hands and objects has gained significant interest in past years
\cite{fan2023arctic,grady2021contactopt,Hasson2020photometric,hasson2021towards,h2odataset,tekin2019ho,yang2021cpf,AbouZeid2023JointTransformer,dexycb,jiang2021hand}.
Methods assume the object template is known - whether rigid~\cite{AbouZeid2023JointTransformer,grady2021contactopt,Hasson2020photometric,hasson2021towards,yang2021cpf,zhu2024grip} or articulated~\cite{fan2023arctic}.
While a majority of the proposed approaches are optimisation-based~\cite{grady2021contactopt,hasson2021towards,zhu2024grip,yang2021cpf, cao2021reconstructing, patel2022learning}, a few are learning-based~\cite{fan2023arctic,AbouZeid2023JointTransformer,Hasson2020photometric,tekin2019ho}.
The closest works to \meth are ArcticNet-SF~\cite{fan2023arctic} and JointTransformer~\cite{AbouZeid2023JointTransformer}, to which we compare.
Similar to \meth, both methods use an encoder-decoder architecture and regress both the hand and object poses. While ArcticNet-SF~\cite{fan2023arctic} uses a ResNet-50~\cite{resnet} backbone, 
JointTransformer replaces it by DINOv2~\cite{oquab2023dinov2} to achieve superior results.
Different from both, \meth leverages hand priors to enrich features and improve results.

Another set of works explore CAD-free object reconstruction~\cite{ye2022s, ye2023diffusion, ye2023ghop, prakash20243d, chen2025hort, huang2022reconstructing, hampali2023hand, fan2024hold, tse2022collaborative,yang2022artiboost}. 
While promising generalisation results are reported, we find that, in practice, the results are not satisfactory; see the \supmat for qualitative comparison with~\cite{fan2024hold,ye2023ghop,ye2023diffusion,sam3dteam2025sam3d3dfyimages}.
Tackling learning-based pose estimation for CAD-free models remains a future direction.

\mypara{Hand Pose Estimation}
Hand pose estimation has been a topic of exploration for decades \cite{ohkawa:ijcv23, cai:eccv18,ge20193d,iqbal2018hand,liu:cvpr24,Mueller2018ganerated,simon2017hand, guo2023handnerf,h2odataset,lee2023im2hands,li2022interacting,rogez2015understanding}. Recently, with large 3D hand pose datasets~\cite{zimmermann2019freihand,jin2020whole,moon:eccv20,fan2023arctic, dexycb, banerjee2024hot3d} and better architectures~\cite{lin2021end,pavlakos2024reconstructing_hamer, Potamias_2025_CVPR_wilor, zhang2025hawor,ye2025predicting}, HaMeR~\cite{pavlakos2024reconstructing_hamer} and WiLoR~\cite{Potamias_2025_CVPR_wilor} scale up model size and training data to achieve robust hand pose estimation on diverse scenes. However, unlike \meth, these methods only target hand pose estimation.
Our key insight is to leverage these strong hand priors for joint hand and object pose estimation.

\mypara{Hand-Object Contact Annotations} 
Most datasets for hand-object reconstruction are
collected in controlled in-lab settings~\cite{fan2023arctic, HOI4D, banerjee2024hot3d,dexycb, yang2022oakink, yu2025dynamic,brahmbhatt2020contactpose,garcia2018first}.
MOW~\cite{cao2021reconstructing} is the only in-the-wild dataset providing paired hand-object meshes.
However, the ground-truth object poses in MOW are only coarsely verified, while the fine-grained contacts between the hand and the object are ignored.
Body pose estimation works, on the other hand, have explored various ways to obtain contact annotations for in-the-wild images.
For example, DECO~\cite{tripathi2023deco} uses a ``vertex painting'' approach to label contact regions on the body. PICO~\cite{cseke_tripathi_2025_pico} extends this by transferring the contact regions to objects.
Motivated by this framework, we propose a pipeline to annotate hand and object contact in egocentric images.

\begin{figure*}[t]
    \centering
    \includegraphics[width=\textwidth]{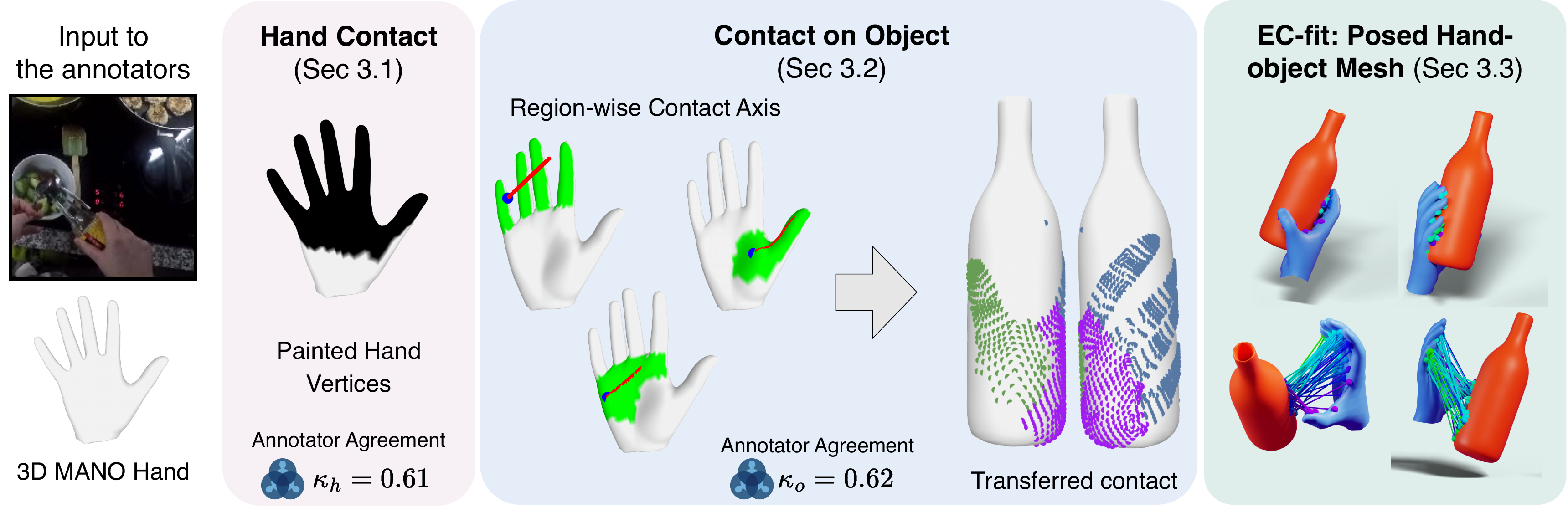}
    \caption{\textbf{\dataset annotation process}.
    Given a hand–object interaction clip, annotators (i) paint contact vertices on a subdivided MANO hand mesh (\cref{section:stageone}); (ii) 
    parametrise each contact region with a 2-DoF contact axis (blue sphere/red line) and transfer it to the object surface with two clicks per axis, yielding bijective hand–object correspondences (\cref{section:stagetwo}); and (iii)
    fit posed hand and object meshes with \fitname (\cref{section:stagethree}).
    The \fitname column visualises the fitted meshes and the one-to-one vertex correspondences. 
    We perform quantitative quality checks on annotations for each stage.
    }
    \label{fig:epic_contact}
\end{figure*}

\section{\dataset Dataset}
\label{section:dataset}

Training and evaluation for hand-object reconstruction require images paired with 3D hand and object pose labels.
We collect new annotations that provide 3D posed hand-object meshes for in-the-wild egocentric images, which we refer to as the \dataset dataset. Importantly, our pipeline includes  a novel approach to manually and efficiently collect bijective 3D contact correspondences.
We then exploit these in an optimisation framework to estimate both hand and object poses.
We describe our annotation pipeline and dataset statistics in this section.

\subsection{Annotating Hand Contact Regions}
\label{section:stageone}
We use videos from the EPIC-Grasps dataset~\cite{zhu2024grip} as it has challenging and diverse hand-object interactions and is paired with 3D meshes for $9$ object classes.
Importantly, the dataset is manually labelled with stable grasp temporal segments, where the hand maintains a steady contact with the object.
This allows annotating a single frame
then applying the same contact across the video.

First step is to annotate regions of the hand that are in contact with the object.
Following previous work~\cite{tripathi2023deco,yang2024egochoir,yang2023lemon}, we label contact regions on 3D meshes using ``vertex painting'' tools.
We annotate the hand mesh because its consistent topology (MANO~\cite{MANO:SIGGRAPHASIA:2017}) simplifies the annotation task and provides a canonical representation, unlike annotating objects with varied and irregular geometries.

We create an interface (see \supmat) where we show the video of a hand in contact with the object and 3D MANO mesh~\cite{MANO:SIGGRAPHASIA:2017}.
Unlike prior work that relies on a single image as input~\cite{tripathi2023deco}, a video clip provides richer context for annotators, particularly as the clip allows multiple views of the contact to be observed during a consistent grasp.
We ask the annotators to ``paint'' contact labels on $N_V = 3106$ vertices of the MANO template mesh, $\mathcal{\tilde{H}} \in \mathbb{R}^{3106\times3}$.
Note that we uniformly subdivide the standard $778$-vertex MANO mesh to $N_V=3106$ vertices, allowing us to capture fine-grained contact regions with high precision.
In case of both hands being in contact, we annotate one hand at a time, specifying the hand side for each annotation.
Furthermore, the annotators are instructed to infer and label all contact regions, including those occluded in the egocentric view, using the video context and grasp motion.
As shown in \cref{fig:epic_contact}, for the bottle in the right hand, vertices are painted on the hand in black.

We validate the accuracy of our annotations using inter-annotator agreement. We calculate the Fleiss' Kappa score ($\kappa_h$) as used in DECO~\cite{tripathi2023deco}.
As shown in \cref{fig:epic_contact}, we report $\kappa_h = 0.61$ across $10$ videos annotated by $12$ annotators (compared to $0.65$ in \cite{tripathi2023deco}).
Additional details on calculation of $\kappa_h$ and a figure showing the highest and lowest $\kappa_h$ examples 
is in the \supmat

\subsection{Contact Regions on Objects}
\label{section:stagetwo}

\mypara{Bijective Contact Transfer} Once we have the contact vertices on the hand, the next step is to get corresponding contact vertices on the object.
We need a bijective mapping between these object vertices and the hand’s contact vertices for obtaining posed hand-object meshes through optimisation.

We follow ContactEdit~\cite{contactedit} and represent the contact regions from \cref{section:stageone} with a contact ``axis''.
This allows us to parametrise each region with a 2-DoF axis, which reduces the problem of mapping the contact region onto the object’s surface to transferring this axis with two clicks: the start of the contact axis and its direction. Using this axis, we can transfer the contact region to the object while preserving the correspondence.
This pipeline is implemented in an interactive web-based tool.

 A key challenge is to identify which contact regions on the hand to parametrise for transfer.
Treating each hand vertex independently is extremely tedious and expensive.
At the other extreme, using a single contact axis for the entire hand is unintuitive and prone to annotation mistakes.
To balance speed, convenience and accuracy, we divide the hand into three regions: the thumb, the four fingers, and the palm (see \cref{fig:epic_contact}). 
We estimate the spacing between fingers using the pose parameters from WiLoR~\cite{Potamias_2025_CVPR_wilor}.
Therefore, for each clip, we have up to three contact axes (depending on contact regions) to transfer, \ie, with at most six clicks, we get the full 3D hand-object contact correspondences.
Further details on the interface are in the \supmat

\mypara{Scaling the Objects} 
The object meshes from~\cite{zhu2024grip} have a standard scale.
However, for transferring the contact region correctly, the scale of the mesh should match that of the object in the image.
We utilise a VLM (Gemini 2.5~\cite{comanici2025gemini25pushingfrontier}) to estimate the scale using a class-specific prompt.
We prompt the VLM to estimate multiple degrees-of-scale for non-uniform scaling, rather than a single isotropic scale factor.
For example, for a ``pan'', we query for both its diameter and its handle length.
This allows us to scale the various parts of the template object mesh to more accurately match the instance in the video.

To verify these scale predictions, we sample $30$ objects covering all $9$ classes and manually compare these to ground truth object sizes.
We identify objects of a known brand (\eg a specific bottle of oil) and measure the dimensions of the same physical object.
This allows us to evaluate the VLM scale estimates against ground truth dimensions, achieving $0.94$ cm MAE ($5.9\%$ relative error) with $82.5\%$ of samples falling within ±$10$\% of the true dimensions.
Additional details on prompts used, degrees-of-scale, and verification analysis are in \supmat

\mypara{Validating the object contact annotations}
We compute the inter-annotator agreement ($\kappa_o$) as $0.62$ on $10$ videos annotated by $4$ annotators.
This confirms that the annotators consistently map hand regions to object surfaces and are able to understand contact despite monocular ambiguities.
Note that $\kappa$ for bijective correspondences on object is not reported in~\cite{tripathi2023deco,cseke_tripathi_2025_pico} for direct comparison.
Additional details along with visualisation are in \supmat

\subsection{\fitname Pipeline: From Contact to Posed Hand-Object Meshes}
\label{section:stagethree}

Finally, we build EPIC-Contact Fitting (\fitname), an optimisation-based pipeline to derive the posed 3D hand-object interaction meshes from contact annotations. 
\fitname aligns the mesh geometry with contact constraints and robustly models the interactions captured in the video space.

\mypara{Pose Initialisation} We initialise the 3D hand pose by applying WiLoR \cite{Potamias_2025_CVPR_wilor} to the central frame $I$ of the clip, yielding a MANO mesh $\mathcal{H}$ with pose $\Theta_{0}$.
For initialising the object pose, we diverge from previous works~\cite{cseke_tripathi_2025_pico,hampali2020honnotate} that utilise standard single-pose initialisation, and find it necessary to
use multiple initialisations,  including random poses and category-aware pose priors from~\cite{zhu2025reconstructing}.

\mypara{Contact-based Alignment} Firstly, we leverage the contact information to guide the object pose alignment with the hand. This is formulated as an optimisation problem. Given the set of bijective vertex pairs $\mathbb{C}:=\{(h_i, o_i)\}$ where $h_i \in \mathcal{H}$ and $o_i \in \mathcal{O}$ are the hand and object vertices in contact, we optimise the object rotation $r_o \in \mathbb{R}^6$ and translation $t_o \in \mathbb{R}^3$ by minimising the contact loss: $\mathcal{L}_{con}=\frac{1}{|\mathbb{C}|}\sum_{i=1}^{|\mathbb{C}|} \left\| h_i - o_i\right\|_2$. 

\mypara{Image-guided Refinement} We use the central frame of the clip ($I$) and an \textbf{occlusion-aware mask loss} to refine the hand and object poses. Traditional approaches~\cite{cseke_tripathi_2025_pico, han2025touch} render the predicted object mesh to a 2D mask $\hat{M}_o$, and directly align it with the object mask $M_o$~\cite{darkhalil2022epic} in the image frame. 
However, masks are incomplete when objects are occluded, which is common in our dataset (\eg plates with food). To this end, we identify an occlusion mask $M_{occ}$ that includes both hand-object occlusions and inter-object occlusions. Our occlusion-aware mask loss only aligns object regions that are not occluded by excluding $M_{occ}$: 
$\mathcal{L}_{m}^{o}=1-IoU([ \hat{M}_o \setminus M_{occ} ], [M_o \setminus M_{occ}])$.  
Moreover, to enforce physical realism, a penetration loss $\mathcal{L}_p$ is added to prevent hand-object interpenetration.

Next, we update the hand pose,  minimising a hand mask loss $\mathcal{L}_m^h$ using the rendered hand mask $\hat{M}_h$ and update the MANO pose $\Theta$. 
A regularisation loss $\mathcal{L}_{reg}=\left\| \Theta - \Theta_{0}\right\|_2$ prevents the refined pose from deviating significantly from the initial pose prediction. We also retain the penetration loss $\mathcal{L}_p$ and contact loss $\mathcal{L}_{con}$.
We manually verify the posed meshes, and correct errors identified by human annotators manually. Details of the manual correction are in \supmat

\begin{figure}[t]
    \centering
    \includegraphics[width=\linewidth]{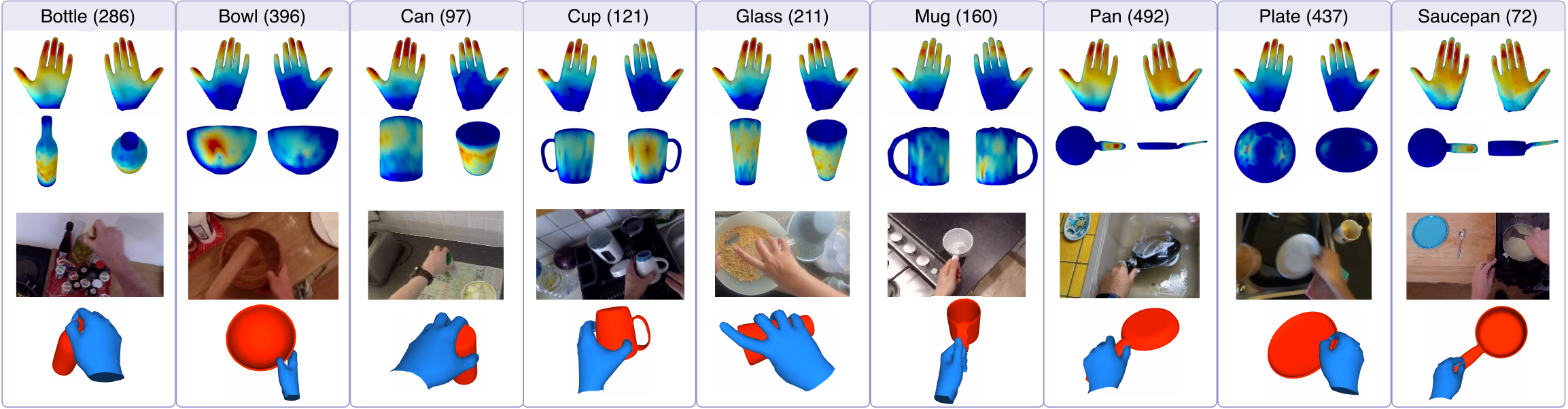}
    \caption{\textbf{EPIC-Contact dataset}. For each object category (\# of annotated clips).
    Middle: contact-frequency heatmaps on the canonical MANO hand mesh and object template mesh.
    Bottom: example frames with \fitname posed hand and object meshes.}
    \label{fig:epiccontact_heatmap}
\end{figure}

\mypara{Clip-level annotations}
So far, we operate at the frame level.
EPIC-Grasps~\cite{zhu2024grip} provides clips with a stable grasp and \fitname yields the object pose relative to the hand in the central frame of the stable grasp,
this allows us to \textit{automatically} broadcast the relative object-to-hand pose across the clip.
We obtain the clip-level object-to-camera poses leveraging hand poses from~\cite{Potamias_2025_CVPR_wilor}.
In case of temporal jitter, frames are labelled with lower confidence.

To summarise, \dataset consists of egocentric interactions with bijective hand–object contact correspondences and posed 3D hand-object meshes, collected via hand contact painting, contact transfer, and \fitname with quality checks. 
\Cref{fig:epiccontact_heatmap} shows the number of samples per object class, heatmaps of aggregate contact on hand and object, showing diversity of grasps as well as a sample of posed hand and object meshes.
In the following sections we describe how we leverage these annotations for training and evaluation.

\section{\meth: \methfull}
\label{method}
\begin{figure*}[t]
    \centering
    \includegraphics[width=\textwidth]{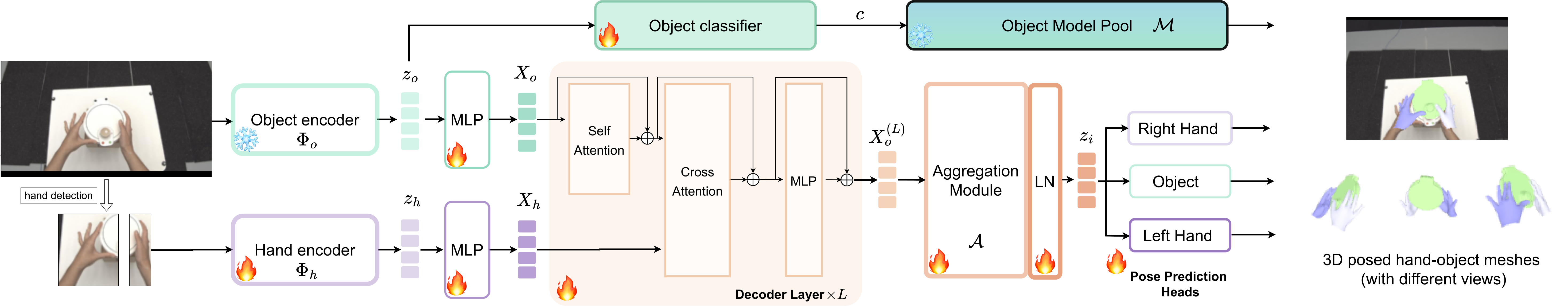}
    \caption{\textbf{Overview of \meth for learning-based hand-object pose estimation.}
    Given an image with hand-object interaction, \meth conditions object features on features from hand pose encoder %
    using an $L$-layer transformer decoder ($\mathcal{D}_{\theta}$).
    The learned features after the aggregation module $\mathcal{A}$ are used to estimate the pose of both hands and interacting object, through dedicated output heads.
    The object head estimates rotation and translation relative to a canonical object pose.
    Furthermore, we predict the object's class $c$ and retrieve the object mesh $\mathcal{O}_c$ from a model pool $\mathcal{M}$.
    The estimated pose is applied to the retrieved object mesh.
    }
    \label{fig:hopformer}
\end{figure*}

In this section, we describe \meth, a learning-based network for estimating hand and object poses.
\Cref{fig:hopformer} provides an overview.
\meth can be trained using either 3D ground truth or posed hand-object meshes obtained from in-the-wild datasets (\cref{section:dataset}).

\label{method:notations}
\subsection{Overview and Notations}

\mypara{Problem Formulation}
Given an image containing one object out of a set of known categories, being manipulated by one or both hands or visible without active manipulation (\eg, resting on a surface), we estimate the pose of each visible hand and the object.
\meth learns to predict hand and object poses across diverse manipulation patterns, including articulated objects when present.

\mypara{Notations}
We denote the $i^{th}$ frame in $k^{th}$ video as $f_k^i$.
We use $r$, $l$, and $o$ to denote the right hand, left hand, and object, respectively.
For hands we use the parametric MANO model~\cite{MANO:SIGGRAPHASIA:2017} to represent the hand's pose and shape by ${\Theta = \{\theta, \beta\}}$.
MANO maps $\Theta$ to a 3D posed and shaped mesh $\mathcal{H}(\Theta) \in \mathbb{R}^{778 \times 3}$, where $\theta\in\mathbb{R}^{48}$ (including global orientation of the hand) and $\beta\in\mathbb{R}^{10}$.
For each object, we predict the object pose
$\omega$, which consists of rotation $\mathbf{R}_o \in \mathbb{R}^6$~\cite{rotation_6d}, translation $\mathbf{T}_o \in \mathbb{R}^3$, and 1D rotation in radians for objects with articulation.
Given this pose $\omega$, we output a posed 3D mesh, $\mathcal{O}_c(\omega) \in \mathbb{R}^{|V_c|\times 3}$ where $|V_c|$ denotes the number of vertices of the object $c$. %

\subsection{Learning-based Pose Estimation}
\label{method:main}

Our goal is a learning-based model for joint hand–object pose estimation, where the estimate of one can inform the other.
During interactions, this context is particularly helpful under occlusion. 

We first extract a generic sequence of object tokens $z_o$ (via $\Phi_o$) and a specialised sequence of hand pose tokens $z_h$ (via $\Phi_h$).
These are linearly projected into a query sequence ($X_o = W_oz_o$) and memory sequence ($X_h = W_h z_h)$, and passed into an $L$-layer Transformer Decoder $\mathcal{D}_{\theta}$.
We use a full decoder stack (rather than a single cross-attention layer) to iteratively refine the object features.
As shown in \cref{fig:hopformer}, in each of the $L$ layers, the object representations: \begin{enumerate*}
    \item communicate via self-attention,
    \item attend to hand context $X_h$ via cross-attention, and
    \item pass through a position-wise feed-forward network.
\end{enumerate*}
This yields a refined object sequence $X_o^{(L)} = \mathcal{D_{\theta}}(X_o, X_h)$.

To obtain the final, aggregated interaction features $z_i$, this output is passed through a residual connection, followed by a learnable linear aggregation module $\mathcal{A}$ and layer normalisation as: 
\label{eq:zi_derivation} 
$z_i = \text{LayerNorm}(\mathcal{A}(X_o^{(L)} + X_o))$.
The resulting $z_i$ is not merely `fused', but is a representation where object features have been progressively modulated by the hand pose.
The core cross-attention mechanism within any layer $\ell$ of our decoder $\mathcal{D}_{\theta}$ is defined as
\begin{equation}
\label{eq:cross_attn}
    \text{CA}(X_o^{(\ell-1)}, X_h) = \delta\left(\frac{(X_o^{(\ell-1)} W_Q^\ell) (X_h W_K^\ell)^\top}{\sqrt{d_k}}\right) (X_h W_V^\ell)
\end{equation}
where $X_o^{(\ell-1)}$ is the object query sequence input to that layer (with $X_o^{(0)} = X_o$), $\delta$ is the Softmax operation, and $W_Q^l$, $W_K^l$, and $W_V^l$ are the learnable projection matrices for the query, key, and value, respectively.

Finally, as shown in \cref{fig:hopformer}, the feature sequence $z_i$ is routed to a set of dedicated prediction heads, which are implemented as small, independent MLPs.
Following the design in~\cite{detr}, different tokens from the sequence $z_i$ are disentangled and learned to predict different components of the hand and object pose.
The first $16$ tokens are used to regress $\theta_r$ (with global orientation), the $17^\text{th}$ token the root, and the $18^\text{th}$ token is used to get the $\beta_r$.
The next $18$ tokens are used to regress pose, orientation ($\theta_l$), root, and shape ($\beta_l$) for the left hand.
The remaining three tokens are used to regress the articulated object pose $\omega$.

\subsection{Object Mesh Retrieval}
\label{section:cad_retrieval}

The primary decoder regresses the object's pose $\omega$ by utilising interaction features, without being hard-coded to a specific object's 3D geometry.
However, to render the 3D hand-object reconstruction, the specific mesh geometry $\mathcal{O}_c$ is required.
We learn to retrieve the correct mesh $\mathcal{O}_c$ from a predefined model pool, $\mathcal{M}$.
We leverage the rich, semantic information encoded in the object features $z_o$ from $\Phi_o$, and add a classification head to predict the object's category, $c$.

The classifier is trained jointly with the pose estimators.
At inference, the predicted class $c$ is used to retrieve the 3D mesh $\mathcal{O}_c$ from $\mathcal{M}$.
The posed mesh is then generated by applying the regressed pose $\omega$ to the retrieved mesh.
This design efficiently reuses features and cleanly decouples the regression of pose from the identification of object class.

\subsection{Training Signal}
\label{section:losses}
To train \meth, we use various frame-level losses.
For hands, we utilise the following losses:
$
    \mathcal{L}_h = \lambda^h_{2D}\mathcal{L}^h_{2D} + \lambda^h_{3D}\mathcal{L}^h_{3D} + \lambda^h_{\theta}\mathcal{L}^h_{\theta} + \lambda^h_{\beta}\mathcal{L}^h_{\beta} + \lambda^h_{T}\mathcal{L}^h_{T}
$
where $h=\{r,l\}$ stands for handedness, $\mathcal{L}_{3D}$ is a supervised loss on the 3D joints (after subtracting the root), $\mathcal{L}^h_{\theta}$ and $\mathcal{L}^h_{\beta}$ are losses on MANO pose and shape parameters, $\mathcal{L}^h_{2D}$ is a 2D loss on the projected 3D points in the image, and $\mathcal{L}^h_{T}$ is the loss on the weak-perspective camera parameters.
Finally, $\lambda^h$ are the weighting coefficients for each of the losses when optimising the hand poses.

Similarly, for the object, there are losses for 3D keypoints $\mathcal{L}^o_{3D}$, 2D projection of 3D keypoints $\mathcal{L}^o_{2D}$, weak-perspective camera parameters, $\mathcal{L}^o_{T}$, classification $\mathcal{L}^o_{c}$, and pose $\mathcal{L}^o_{\omega}$ on the object:
$
    \mathcal{L}_o = \lambda^o_{3D}\mathcal{L}^o_{3D} + \lambda^o_{2D}\mathcal{L}^o_{2D} + \lambda^o_{T}\mathcal{L}^o_{T} + \lambda^o_{c}\mathcal{L}^o_{c} + \lambda^o_{\omega}\mathcal{L}^o_{\omega}.
$

For frames where the hand is in contact with an object, we add a CDev-based interaction loss (details in \cref{section:metric}):
$
    \mathcal{L}_{int} = \lambda^{int}_{ro}\mathcal{L}^{int}_{ro} + \lambda^{int}_{lo}\mathcal{L}^{int}_{lo}.
$
The final loss term for training \meth is then:
$
    \mathcal{L} = \mathcal{L}_r + \mathcal{L}_l + \mathcal{L}_o + \mathcal{L}_{int}.
$
The losses above use MSE except for $\mathbf{R}_o$ in $\mathcal{L}^o_{\omega}$ which uses geodesic loss.

\section{Experiments and Results}
\label{experiments}

\subsection{Datasets}
\mypara{ARCTIC~\cite{fan2023arctic}} 
This in-the-lab dataset is captured in a constrained setting where a subject is recorded manipulating one object.
It consists of $2.1$M RGB frames across $9$ camera views where $10$ participants manipulate $11$ objects with both hands.
Participants either ``use'' or ``grasp'' the object during the interactions.
The poses of the hands and objects are captured using a MoCap system.
ARCTIC offers $8$ exocentric views for pre-training and $1$ egocentric view to fine-tune and evaluate.
We follow the original train and validation splits~\cite{fan2023arctic}.
As this dataset uses MoCap ground-truth, we use it for ablation experiments.

\mypara{\dataset}
In \dataset (which we collect following the details in ~\cref{section:dataset}) we have $2{,}272$ videos consisting of $62.3$K frames across $9$ objects.
Note that the same posed hand-object mesh is used for training all frames within the same video clip, as these clips are temporally annotated as a continuous stable grasp~\cite{zhu2024grip}.
We use $237$ videos for testing and $2{,}035$ videos as training, keeping the distribution of hand side and object categories identical between the sets.

\subsection{Implementation Details}
For all experiments, we use DINOv2~\cite{oquab2023dinov2} with ViT-G backbone~\cite{vit} as our object feature extractor ($\Phi_o$) and WiLoR~\cite{Potamias_2025_CVPR_wilor} as $\Phi_h$ for hand features.
The decoder depth $L$ is set to $12$.
The hyper-parameters $\lambda$s in the loss terms are all set to $1$ except for $\lambda^h_{2D} = 5.0$, $\lambda^r_{3D} = \lambda^o_{3D} = 5.0$, $\lambda_{\theta}^h = \lambda^l_{3D} = 10.0$ and $\lambda^h_{\beta} = \lambda^o_{c} = 0.001$.

We train using the AdamW~\cite{Loshchilov2017DecoupledWD} optimiser.
Following~\cite{fan2023arctic}, \meth is trained in two stages, the first stage is trained on exocentric views and then fine-tuned on egocentric views from ARCTIC.
While training on exocentric views, we use a linear warm-up of the learning rate, from $1$e-$7$ to $5$e-$5$ in first $5\%$ of steps, and cosine decay from $5$e-$5$ to $1$e-$7$ for the remaining steps.
The batch size is $256$ across $4$ NVIDIA GH200 GPUs and we train for $25$ epochs.
For egocentric training, we use a learning rate of $3$e-$5$ with cosine decay from $1$e-$7$, batch size of $128$ across $4$ GPUs, trained for $30$ epochs with early stopping.
The same hyper-parameters are used for training on \dataset for $125$ epochs with early stopping.
We use a weak perspective camera model~\cite{fan2023arctic,boukhayma20193d,kanazawaHMR18,Kocabas_PARE_2021,9008830} for translation and a 6D representation for rotation~\cite{rotation_6d}.

\subsection{Quantitative Metrics}
\label{section:metric}

We follow~\cite{fan2023arctic} and report metrics capturing contact/relative pose, motion, hand accuracy, and object pose estimation.
\begin{enumerate*}
    \item \textbf{Contact Deviation (CDev, mm)}: The mean distance between corresponding hand--object contact vertex pairs;
    \item \textbf{Mean Relative-Root Position Error (MRRPE$_{rl/ro}$, mm)}: The relative root translation error for hand--hand and hand--object;
    \item \textbf{Motion Deviation (MDev, mm)}: Measuring disagreement in motion of vertices in stable contact across consecutive frames;
    \item \textbf{Acceleration Error (ACC$_{h/o}$, m/s$^2$)}: Measuring smoothness via acceleration differences for hand/object vertices;
    \item \textbf{Mean Per-Joint Position Error (MPJPE, mm)}: The mean 3D error over 21 hand keypoints;
    \item \textbf{Average Articulation Error (AAE, $^\circ$)}: Articulation error for articulated objects;
    \item \textbf{Success Rate (SR@0.05/SR@0.1, \%)}: The fraction of object vertices within 5\%/10\% of the object diameter;
    \item \textbf{Object Classification Accuracy (Cls)}: The accuracy of the object classification head. 
\end{enumerate*}
For \dataset, where symmetric objects are used (\eg bottle, can), CDev, MDev, ACC and Success Rate are updated to be symmetry aware.
Details in \supmat

\subsection{Baselines}

We use learning-based baselines closest to \meth. 
\begin{enumerate*}
    \item \textbf{ArcticNet-SF}~\cite{fan2023arctic}: Encoder-decoder architecture with ResNet-50~\cite{resnet} as its backbone.
    \item \textbf{JointTransformer}~\cite{AbouZeid2023JointTransformer}: Builds on ArcticNet-SF and replaces the CNN-based backbone with DINOv2~\cite{oquab2023dinov2}.
    As this uses the same feature backbone as us, it also serves as the best baseline for direct comparison. Improvements over this baseline are results of our proposed \meth design choices.
    Importantly, JointTransformer remains the current SOTA method on the ARCTIC dataset reconstruction leaderboard.
\end{enumerate*}
For fair comparison, we retrain both methods on ARCTIC and \dataset using the same splits and evaluation protocol as ours.

\subsection{Results}

\begin{table*}[t]
    \centering
    \caption{\textbf{ARCTIC (Egocentric; in-lab)}.
    \meth outperforms baselines by a clear margin on all metrics (except AAE).
    \textbf{Bold} numbers are best performance.
    }
    \begin{adjustbox}{width=\textwidth}
    \begin{tabular}{@{}lcccccccc@{}}
        \toprule
        \multirow{2}{*}{Method} & \multicolumn{2}{c}{Contact and Relative Positions} & \multicolumn{2}{c}{Motion} & Hand & \multicolumn{3}{c}{Object} \\
        \cmidrule(lr){2-3}\cmidrule(lr){4-5}\cmidrule(lr){6-6}\cmidrule(l){7-9}
        & CDev [$mm$] ↓ & MRRPE$_{rl/ro}$ [$mm$] ↓ & MDev [$mm$] ↓ & ACC$_{h/o}$ [$m/s^2$] ↓ & MPJPE [$mm$] ↓ & AAE [$^{\circ}$] ↓ & SR@0.05 [\%] ↑ & Cls [\%] ↑ \\
        \midrule
        ArcticNet-SF~\cite{fan2023arctic} & $44.1$ & $33.9$ / $36.8$ & $11.8$ & $6.3$ / $11.3$ & $22.9$ & $8.0$ & $59.0$ & - \\
        JointTransformer~\cite{AbouZeid2023JointTransformer} & $35.0$ & $34.0$ / $29.9$ & $10.4$ & $6.9$ / $10.1$ & $20.0$ & $\mathbf{4.9}$ & $76.2$ & - \\
        \meth (\textit{ours}) & $\mathbf{31.9}$ & $\mathbf{31.1}$ / $\mathbf{29.4}$ & $\mathbf{7.3}$ & $\mathbf{4.8}$ / $\mathbf{6.2}$ & $\mathbf{16.1}$  & $5.0$ & $\mathbf{82.4}$ & $\mathbf{99.5}$ \\
        \bottomrule
    \end{tabular}
    \end{adjustbox}
    \label{tab:ego_arctic_results}
\end{table*}

\begin{table*}[t]
    \centering
    \caption{\textbf{EPIC-Contact (Egocentric; in-the-wild).} \meth improves interaction and object pose estimation over all baselines.
    \textbf{Bold} numbers are best performance.
    MRRPE$_{rl}$ is invalid for \dataset due to one hand per sequence.
    }
    \begin{adjustbox}{width=\textwidth}
    \begin{tabular}{@{}lcccccccc@{}}
        \toprule
        \multirow{2}{*}{Method} & \multicolumn{2}{c}{Contact and Relative Positions} & \multicolumn{2}{c}{Motion} & Hand & \multicolumn{3}{c}{Object} \\
        \cmidrule(lr){2-3}\cmidrule(lr){4-5}\cmidrule(lr){6-6}\cmidrule(l){7-9}
        & CDev [$mm$] ↓ & MRRPE$_{ro}$ [$mm$] ↓ & MDev [$mm$] ↓ & ACC$_{h/o}$ [$m/s^2$] ↓ & MPJPE [$mm$] ↓ & SR@0.05 [\%] ↑ & SR@0.1 [\%] ↑ & Cls [\%] ↑ \\
        \midrule
         ArcticNet-SF~\cite{fan2023arctic} & $94.2$ & $166.7$ & $70.1$ & $3.9$ / $5.3$ & $44.8$ & $16.5$ & $55.9$ & - \\
        JointTransformer~\cite{AbouZeid2023JointTransformer} & $30.1$ & $78.6$ & $20.0$ & $3.1$ / $9.5$ & $22.9$ & $17.6$ & $56.9$ & - \\
        \meth (\textit{ours}) & $\mathbf{20.7}$ & $\mathbf{65.8}$ & $\mathbf{11.4}$ & $\mathbf{2.5}$ / $\mathbf{4.1}$ & $\mathbf{19.9}$ & $\mathbf{29.8}$ & $\mathbf{69.7}$ & $\mathbf{52.9}$ \\
        \bottomrule
    \end{tabular}
    \end{adjustbox}
    \label{tab:ego_epic_results}
\end{table*}

\mypara{In-lab ARCTIC Results}
As shown in \cref{tab:ego_arctic_results}, \meth achieves the best performance on ARCTIC’s egocentric split, improving contact consistency, motion, and object pose estimation over prior learning-based methods.
Compared to JointTransformer~\cite{AbouZeid2023JointTransformer}, we reduce CDev from 35.0$\rightarrow$31.9\,mm and MDev from 10.4$\rightarrow$7.3\,mm, while improving SR@0.05 from 76.2$\rightarrow$82.4; hand accuracy also improves substantially (MPJPE 20.0$\rightarrow$16.1\,mm), with comparable articulation error (AAE 4.9$\rightarrow$5.0).
We also present results on the exocentric split in \supmat, where \meth outperforms JointTransformer by an equal or larger margin on most metrics.
These gains establish \meth as the new SOTA on this established benchmark and support our central design choice of conditioning object features on strong hand priors for robust hand--object pose estimation.

\mypara{In-the-wild \dataset Results}
\Cref{tab:ego_epic_results} shows that \meth improves pose-estimation and interaction consistency on \dataset, outperforming learning-based baselines under in-the-wild occlusions and clutter.
Compared to JointTransformer~\cite{AbouZeid2023JointTransformer}, \meth reduces CDev from $30.1\rightarrow20.7$\,mm and MDev from $20.0\rightarrow11.4$\,mm, while improving SR@0.05 from $17.6\rightarrow29.8$; hand accuracy also improves (MPJPE $22.9\rightarrow19.9$\,mm).
We additionally predict object category to enable fully automatic inference; however, to match prior work that assumes oracle object meshes, we report pose with oracle mesh and provide fully automatic results in the \supmat

These results also highlight the additional challenge of \dataset compared to established benchmarks, as expected. For example, JointTransformer drops SR@0.05 from $76.2\%$ on ARCTIC (Ego) to $17.6\%$ on EPIC-Contact.
This is not unexpected given ARCTIC shares the same object instances between train and val split, while \dataset shows novel instances of the known object categories, often transparent, occluded and in a cluttered scene.

\begin{table*}[!t]
    \centering
    \caption{
    \textbf{Architectural and loss ablations on ARCTIC (egocentric).}
We ablate \meth components and training objectives.
Results show that hand-conditioned cross-attention decoding 
and the interaction/object supervision terms are key to performance across pose-estimation, contact, and motion metrics.}
    \begin{adjustbox}{width=\textwidth}
\begin{tabular}{@{}lccccccc@{}}
    \toprule
    \multirow{2}{*}{Setting} & \multicolumn{2}{c}{Contact and Relative Positions} & \multicolumn{2}{c}{Motion} & Hand & \multicolumn{2}{c}{Object} \\
    \cmidrule(lr){2-3}\cmidrule(lr){4-5}\cmidrule(lr){6-6}\cmidrule(l){7-8}
    & CDev [$mm$] ↓ & MRRPE$_{rl/ro}$ [$mm$] ↓ & MDev [$mm$] ↓ & ACC$_{h/o}$ [$m/s^2$] ↓ & MPJPE [$mm$] ↓ & AAE [$^{\circ}$] ↓ & SR@0.05 [\%] ↑ \\
    \midrule

    \multicolumn{8}{@{}l}{\textbf{Loss ablations}} \\
    \addlinespace[2pt]
    \textit{w/o} Interaction Loss ($\mathcal{L}_{int}$) & $40.7$ & $33.3$ / $36.5$ & $10.8$ & $6.2$ / $8.4$ & $18.3$ & $7.7$ & $79.2$ \\
    \textit{w/o} 3D keypoints ($\mathcal{L}_{3D}^{o,h}$) & $36.8$ & $33.7$ / $33.8$ & $10.0$ & $6.1$ / $8.7$ & $18.7$ & $7.8$ & $78.5$ \\
    \textit{w/o} 2D keypoints ($\mathcal{L}_{2D}^{o,h}$) & $35.6$ & $35.9$ / $33.0$ & $9.5$ & $5.9$ / $8.1$ & $18.3$ & $7.4$ & $79.5$ \\
    \textit{w/o} Beta ($\mathcal{L}_{\beta}^h$) & $35.4$ & $33.2$ / $32.6$ & $9.6$ & $5.8$ / $8.4$ & $18.1$ & $7.2$ & $79.3$ \\
    \textit{w/o} Translation ($\mathcal{L}_T^{o,h}$) & $35.0$ & $32.8$ / $31.5$ & $9.6$ & $5.9$ / $8.3$ & $18.0$ & $7.2$ & $79.6$ \\
    \textit{w/o} Object Pose ($\mathcal{L}_\omega^o$) & $34.5$ & $31.7$ / $31.0$ & $9.8$ & $5.7$ / $9.5$ & $17.3$ & $7.4$ & $72.5$ \\
    \textit{w/o} Object Losses ($\mathcal{L}_o$) & $33.8$ & $31.3$ / $31.4$ & $9.5$ & $5.6$ / $9.4$ & $17.3$ & $8.6$ & $66.8$ \\

    \addlinespace[3pt]
    \midrule
    \addlinespace[1pt]
    \multicolumn{8}{@{}l}{\textbf{Architecture ablations}} \\
    \addlinespace[2pt]
    Concat + MLP & $74.4$ & $48.1$ / $66.5$ & $32.0$ & $6.6$ / $25.5$ & $19.6$ & $18.7$ & $29.4$ \\
    DINOv2 as hand features & $44.2$ & $36.8$ / $37.1$ & $16.2$ & $7.4$ / $13.0$ & $20.6$ & $6.3$ & $70.0$ \\
    No SA & $61.3$ & $74.0$ / $60.4$ & $17.9$ & $7.3$ / $14.4$ & $22.9$ & $10.9$ & $55.8$ \\
    No Aggr. Module ($\mathcal{A}$) & $40.4$ & $35.9$ / $36.6$ & $14.0$ & $6.0$ / $11.4$ & $16.8$ & $6.3$ & $74.0$ \\
    $\mathcal{L}=1$ & $101.6$ & $93.5$ / $90.8$ & $34.9$ & $15.0$ / $25.4$ & $46.7$ & $23.4$ & $21.9$ \\
    $\mathcal{L}=6$ & $59.2$ & $54.0$ / $49.4$ & $19.5$ & $9.8$ / $15.2$ & $24.0$ & $8.2$ & $58.6$ \\
    $\mathcal{L}=9$ & $50.0$ & $47.1$ / $42.8$ & $15.7$ & $6.8$ / $11.5$ & $21.5$ & $7.9$ & $69.0$ \\
    
    \addlinespace[3pt]
    \midrule
    \addlinespace[1pt]
    \multicolumn{8}{@{}l}{\textbf{Main result}} \\
    \addlinespace[2pt]
    \rowcolor{gray!8}
    \meth (\textit{ours}) & $\mathbf{31.9}$ & $\mathbf{31.1}$ / $\mathbf{29.4}$ & $\mathbf{7.3}$ & $\mathbf{4.8}$ / $\mathbf{6.2}$ & $\mathbf{16.1}$ & $\mathbf{5.0}$ & $\mathbf{82.4}$ \\
    \bottomrule
\end{tabular}
    \end{adjustbox}
    \label{tab:loss_ablation}
\end{table*}

\begin{figure}[t]
    \centering
    \includegraphics[width=\linewidth]{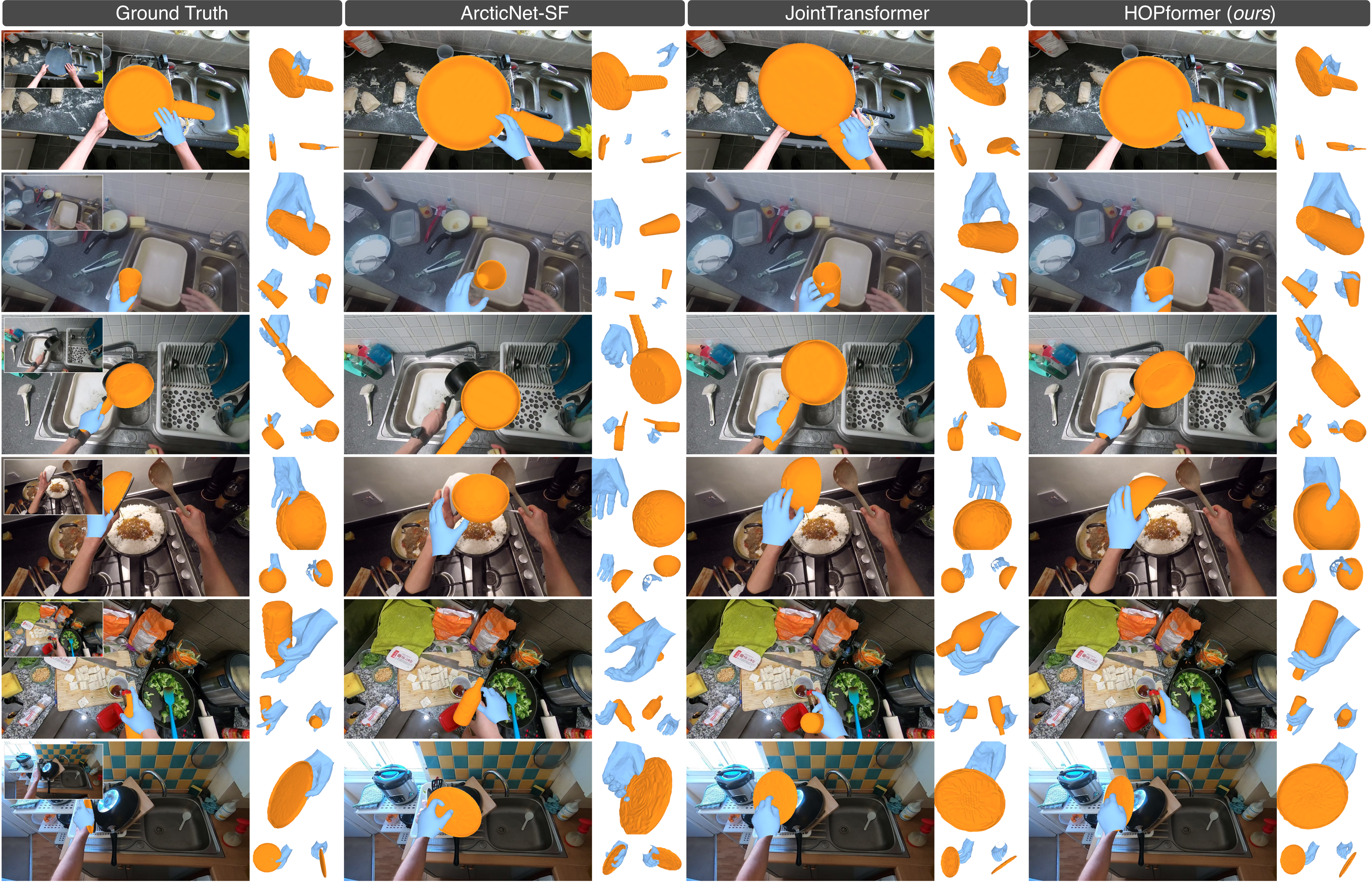}
    \caption{\textbf{Qualitative comparison with baselines.} 
    Original image (thumbnail), image with projected hand-object mesh, and three views of the posed meshes.
    \meth has visibly superior estimations of pose compared to the baselines.
    }
    \label{fig:arctic-quali}
\end{figure}

\mypara{Qualitative Results}
We demonstrate qualitative comparison of models' predictions on \dataset in \cref{fig:arctic-quali}.
\meth estimates plausible hand and object pose under heavy occlusion, for diverse objects in challenging scenes. Prior works fail in most cases, even in securing contact between the hand and the object.
Failure case (bottom) is impacted by strong priors from learnt data, where a plate is expected to be flat.
Additional results are provided in the \supmat

\subsection{Ablations}

All ablations are reported on the ARCTIC dataset consisting of 3D MoCap ground truth (egocentric split).

\mypara{Ablating the losses}
In \cref{tab:loss_ablation}, we ablate each training term and observe consistent drops, with the largest effects coming from object supervision and interaction constraints.
Removing the object losses ($\mathcal{L}_o$) reduces SR@0.05 from $82.4\rightarrow66.8$ and increases AAE from $5.0\rightarrow8.6$, while removing the object pose term ($\mathcal{L}^o_{\omega}$) also notably reduces SR ($82.4\rightarrow72.5$).
Removing the CDev-based interaction loss ($\mathcal{L}_{int}$) primarily hurts physical consistency, increasing CDev from $31.9\rightarrow40.7$ and MDev from $7.3\rightarrow10.8$.
For hand accuracy, keypoint supervision is most critical: removing $\mathcal{L}^{3D}_{o,h}$ yields the largest MPJPE increase ($16.1\rightarrow18.7$), with $\mathcal{L}^{2D}_{o,h}$ and $\mathcal{L}_{int}$ showing similar degradation ($\rightarrow18.3$), whereas $\mathcal{L}^h_{\beta}$ and $\mathcal{L}^{T}_{o,h}$ provide smaller but consistent gains.

\mypara{Architectural Ablations}
Also in \cref{tab:loss_ablation}, we ablate the architectural design of \meth.
\textbf{Concat + MLP}. 
Replacing the Cross-Attention decoder with naive fusion causes a sharp collapse (SR $82.4\rightarrow29.4$, CDev $31.9\rightarrow74.4$), despite using the same strong backbones. This confirms that HOPformer’s gains come from conditioning features and not from the backbone.
\textbf{DINOv2 as hand features}. 
Using generic DINOv2 tokens instead of WiLoR weakens the hand prior (MPJPE $16.1\rightarrow20.6$) and propagates to object quality (SR $82.4\rightarrow70.0$). 
This highlights that pose-specialised hand features are key to effective hand-conditioned object refinement.
\textbf{No self-attention}. 
Disabling self-attention, which enriches the object features, negatively impacts token communication, leading to large degradations (SR $82.4\rightarrow55.8$, MRRPE $31.1$/$29.4\rightarrow74.0$/$60.4$). 
This indicates that token-to-token interaction is critical for accurate object estimation.
\textbf{No Aggregation Module.}
Removing the learnable aggregation module, and instead using a fixed subset of the tokens for each output head, 
consistently hurts performance (SR $82.4\rightarrow74.0$, CDev $31.9\rightarrow40.4$). 
This indicates that 
\meth benefits from learnt pooling into a compact interaction representation.
\textbf{Varying decoder depth.} 
Reducing decoder depth sharply reduces performance ($L=1$: SR $21.9$), while increasing depth steadily improves it up to $L=12$ (SR $82.4$). 
This supports our design choice of an $L$-layer decoder for iterative refinement of object tokens under hand guidance.

Overall, these ablations confirm that HOPformer’s improvements are obtained by iterative, hand-conditioned object refinement.

\begin{figure}
    \centering
    \begin{minipage}[t]{0.55\linewidth}
        \vspace{0pt} %
        \mypara{Scaling the number of object categories}
        To evaluate \meth's performance as the number of objects increases, we train \meth on $N \in \{3, 6, 9, 11\}$ object classes from ARCTIC.
        \cref{fig:num_classes_ablation} shows a line plot with MDev(ho) score as number of objects increases.
        \meth consistently benefits as more object categories are added highlighting superior generalisability despite the increase in the challenge.
        Other metrics show similar gains and are in the \supmat

    \end{minipage}\hfill
    \begin{minipage}[t]{0.4\linewidth}
        \vspace{0pt} %
        \centering
        \includegraphics[width=\linewidth]{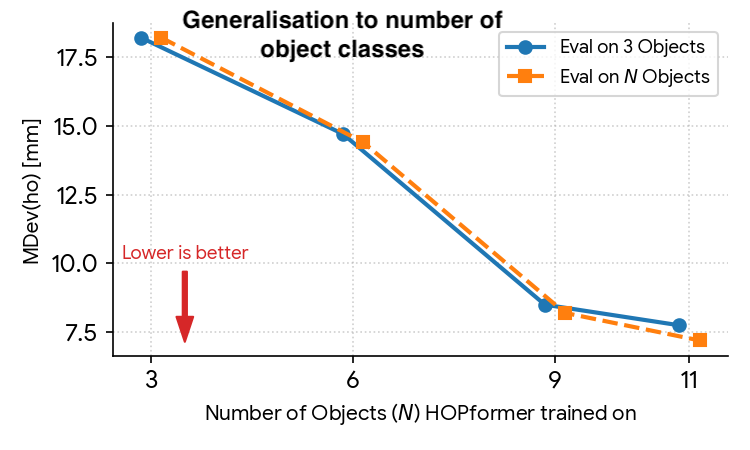}
        \caption{MDev(ho) improves as the number of objects increases.}
        \label{fig:num_classes_ablation}
    \end{minipage}
\end{figure}

\mypara{Training \meth from scratch on \dataset}
\Cref{tab:epic_from_scratch} shows initialising \meth with ARCTIC weights yields consistent gains on \dataset,  notably in relative pose and motion (MRRPE$_{ro}$ $99.5\rightarrow65.8$, MDev $19.5\rightarrow11.4$).
\meth benefits from additional data during the pre-training.

\noindent\textbf{Compute/Runtime:} On one NVIDIA GH200, with batch=1 for online evaluation, HOPformer runs in $106.63$\,ms median per frame at batch=$1$ ($9.12$ frame/s); additional compute and profiling details are provided in the \supmat

\begin{table*}[t]
    \centering
    \caption{\textbf{Cross-dataset transfer (ARCTIC $\rightarrow$ \dataset)}.
Training \meth from scratch on EPIC vs.\ ARCTIC-initialised fine-tuning.}
    \begin{adjustbox}{width=\textwidth}
    \begin{tabular}{@{}lccccccc@{}}
        \toprule
        \multirow{2}{*}{Method} & \multicolumn{2}{c}{Contact and Relative Positions} & \multicolumn{2}{c}{Motion} & Hand & \multicolumn{2}{c}{Object} \\
        \cmidrule(lr){2-3}\cmidrule(lr){4-5}\cmidrule(lr){6-6}\cmidrule(l){7-8}
        & CDev [$mm$] ↓ & MRRPE$_{ro}$ [$mm$] ↓ & MDev [$mm$] ↓ & ACC$_{h/o}$ [$m/s^2$] ↓ & MPJPE [$mm$] ↓ & SR@0.05 [\%] ↑ & SR@0.1 [\%] ↑ \\
        \midrule
         \meth (scratch) & $24.4$ & $99.5$ & $19.5$ & $3.5$ / $7.1$ & $54.8$ & $3.9$ & $24.2$ \\
        \meth (ARCTIC init; \textit{ours}) & $\mathbf{20.7}$ & $\mathbf{65.8}$ & $\mathbf{11.4}$ & $\mathbf{2.5}$ / $\mathbf{4.1}$ & $\mathbf{19.9}$ & $\mathbf{29.8}$ & $\mathbf{69.7}$ \\
        \bottomrule
    \end{tabular}
    \end{adjustbox}
    \label{tab:epic_from_scratch}
\end{table*}
\section{Conclusion}

This paper contributes to 3D hand-object pose estimation for in-the-wild egocentric images in two ways.

First, we enable training and evaluation in-the-wild by annotating and releasing the \dataset dataset.
We annotate \dataset video clips with hand contact regions and bijective contact on $2.3$K egocentric clips with functional interactions involving $9$ object classes.
\dataset is quality checked, diverse and a significantly challenging benchmark to guide future research.

Second, we propose \meth \  -- a learning-based approach that, given an RGB image, predicts the pose of hands and object in a single forward pass.
\meth uses hand priors that enrich the representation, providing superior results in joint estimation of hand, object class, and object pose. 
We outperform SOTA on the established ARCTIC benchmark and provide a strong starting point for evaluation on EPIC-Contact.

\section*{Acknowledgement}

This work was supported by EPSRC Programme Grant Visual AI (EP/T028572/1) and EPSRC Fellowship UMPIRE (EP/T004991/1). 
S Bansal is supported by a Charitable Donation to UoB from Meta.
Z Zhu and J Zhao are supported by UoB-CSC Scholarships.
While MJB is employed by Epic Games, this work was performed solely at, and funded solely by, the Max Planck Society.
We acknowledge the usage of GPU Node
hours granted as part of the  AIRR Innovator project ``5D Hand-Object Interaction Modelling from In-the-wild Videos'' (Mar 2026 - Sep 2026), AIRR Gateway project ``HOI Foundational Model from Egocentric Data'' (Dec 2025 - Mar 2026) and the Sovereign AI Unit call project
``Gen Model in Ego-sensed World'' (Aug 2025 - Nov 2025). 

The authors would like to thank Jacob Chalk, Tomoya Yoshida, Kranti Kumar Parida, Omar Emara, and Balamurugan Thambiraja for their comments on the manuscript.
We thank Rajan, Saikiran, Durga Prasad and their team from Elancer for assisting with annotating the \dataset dataset.

\ifSubfilesClassLoaded{%
  \bibliographystyle{splncs04}
  \bibliography{main}
}{}
\end{document}

\bibliographystyle{splncs04}
\bibliography{main}
\clearpage

\title{Supplementary:\\Towards in-the-wild Egocentric 3D Hand-Object Pose Estimation}

\newpage

\section*{Appendix}

This appendix provides supplementary information for the main paper.
\Cref{sec:additional_details_hopformer} provides additional details on \meth including the compute, metrics, implementation details, qualitative results, scalability, comparison to CAD-free methods, and exocentric results on ARCTIC.
\Cref{sec:our_dataset} provides additional information on the EPIC-Contact dataset.
\Cref{sec:additional_relevant_works} discusses other relevant works.
\Cref{sec:limitation_and_future} discusses limitations and future directions.
\Cref{section:scale_prompt} provides the prompts used for obtaining the scales of the objects.

\section{Additional Details on \meth}
\label{sec:additional_details_hopformer}

\subsection{Compute time analysis}
\chg{All results were obtained on a single NVIDIA GH200 120\,GB GPU.
We report statistics for a single-sample forward pass using an end-to-end wall-clock timer with explicit CUDA synchronisation.
The input sample was preloaded onto the GPU to exclude dataloader overhead.
We used 30 warmup iterations followed by 300 timed iterations; latency is summarised with P50 (median) and P95.
\Cref{tab:runtime} summarises the results, the median time taken per sample is $106.63$ ms.
Compared to >$10$s for optimisation-based methods~\cite{hasson2021towards,zhu2024grip} \meth performs feed-forward inference and is thus substantially faster (in milliseconds).
\meth utilises 
$1.157\times 10^{12}$ FLOPs per forward ($\approx 1157$ GFLOPs).}

\begin{table}[h!]
\centering
\caption{Compute and runtime for a single-sample forward pass. Timings are end-to-end wall-clock with explicit CUDA synchronisation; 30 warmup + 300 timed iterations; input batch preloaded to exclude dataloader overhead.}
\setlength{\tabcolsep}{6pt}
\begin{adjustbox}{width=\textwidth}
\begin{tabular}{@{}lllllll@{}}
\toprule
Hardware & Params & FLOPs / sample & Peak mem (alloc) & Latency P50 & Latency P95 & Throughput \\
GH200 120\,GB & 1.83B & 1.157 TFLOPs & 7.01 GiB & 106.63 ms & 165.89 ms & 9.12 samp/s \\
\bottomrule
\end{tabular}
\end{adjustbox}
\label{tab:runtime}
\end{table}

\subsection{Details on symmetry-aware metrics}

In \cref{section:metric},
we define the quantitative metrics used for the evaluation (originally in~\cite{fan2023arctic}) and note that, for \dataset, CDev, MDev, ACC, and Success Rate are updated to be symmetry aware due to presence of symmetric objects.
Here, we provide additional details of these symmetry-aware variants of the evaluation metrics.

The goal is to preserve the original evaluation protocol while avoiding penalising predictions that are correct up to symmetry.
This is critical for object classes in EPIC-Contact that are symmetric (e.g. bottle).
Note that changing CDev and SR to be symmetry-aware makes them invariant to sliding along the object surface, as the symmetry-aware variants become invariant not only to the object's rotational symmetry.
In contrast, MRRPE and MDev penalise sliding even when symmetry-aware.

\mypara{Contact Deviation (CDev, mm)}
In the standard definition, CDev measures the mean distance between corresponding hand-object contact vertex pairs.
For symmetric objects, however, a prediction may place contact on an equivalent object location without matching the annotated object vertex exactly.
We therefore use the ground-truth contact distances to identify the contact  hand vertices, and for these vertices measure the distance to the closest vertex on the predicted object mesh.

\mypara{Motion Deviation (MDev, mm)}
We keep the same stable contact windows defined from the ground-truth contact annotations, but do not require the predicted motion to follow the exact annotated object vertex throughout the window.
Instead, object motion is measured using a fixed predicted object patch associated with the in-contact hand vertex at the start of the window, and MDev is computed as the disagreement between the hand and object motion across consecutive frames.
This makes the metric robust when object locations are equivalent under symmetry but penalises when an object slips (i.e. changes the in-contact vertices over time).

\mypara{Acceleration Error (ACC$_{h/o}$, m/s$^2$)}
We do not make any changes to ACC$_h$.
We update ACC$_o$ to be symmetry aware, for symmetric objects.
After removing the object root, we compute acceleration differences for object vertices and compare prediction and ground truth after matching each object vertex to the closest equivalent vertex in 3D.
ACC$_o$ is then averaged in both directions, so that the metric continues to measure motion smoothness while avoiding penalising equivalent vertex assignments on symmetric objects.

\mypara{Success Rate (SR@0.05/SR@0.1, \%)}
Rather than evaluating object vertices with fixed vertex correspondences, we compare each predicted object vertex to the closest ground-truth object vertex after removing the object root.
Success Rate is then computed as the fraction of object vertices within 5\%/10\% of the object diameter.
This preserves the same diameter-normalised criterion used in the paper while avoiding penalising predictions that are correct up to symmetry.

\subsection{Additional Implementation Details}

The feature dimension from the object encoder, DINOv2~\cite{oquab2023dinov2} ($\Phi_o$) is $1536$ and that from the hand encoder WiLoR~\cite{Potamias_2025_CVPR_wilor} ($\Phi_h$) is $1280$.
We use an MLP to project features of $\Phi_h$ to object's embedding space.
The output of the $L$-layer decoder $X_o^{(L)}$ is added with object features ($X_o$) and then passed through the Aggregation module ($\mathcal{A}$).
Aggregation module is an MLP layer that reduces the number of tokens from $256$ to $39$ required for regression of object and hands' pose (described in~\cref{{method:main}}).

When training on ARCTIC, data augmentation is applied to the images, scaling ($\pm 25\%$), color jitter ($\pm40\%$), and rotation ($\pm30\%)$.
Following~\cite{fan2023arctic}, for the predicted weak perspective camera we use a fixed focal length of $1000.0$ for ARCTIC and of $5000.0$ for \dataset.
These fixed focal lengths are used to obtain translation for hands and object in the scene.

\subsection{Qualitative results}

\begin{figure*}
    \centering
    \includegraphics[width=\linewidth]{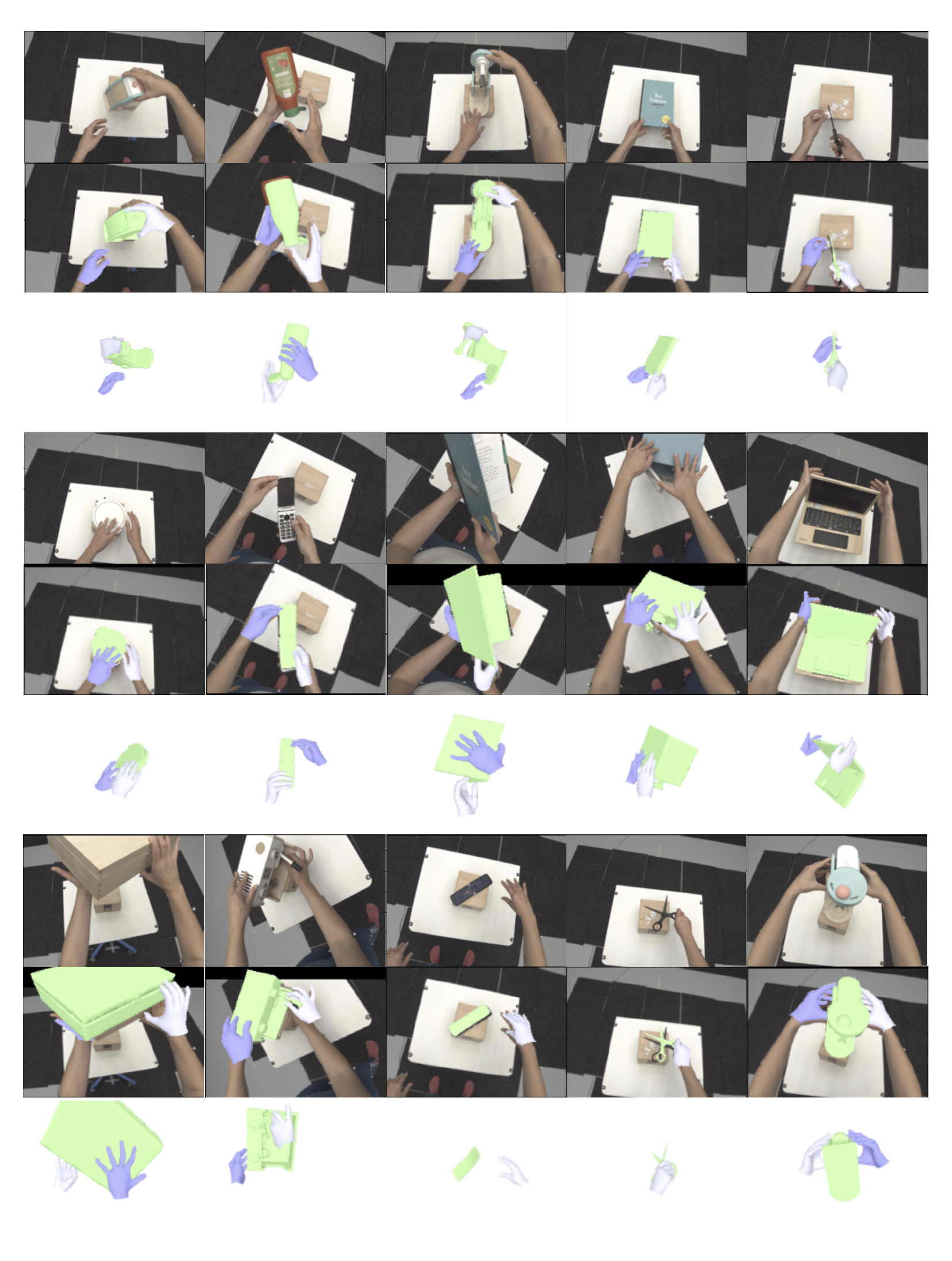}
    \caption{\textbf{ARCTIC Qualitative Results.}
    \meth performs well for cases with both hands or with one hand.
    For small objects like scissors and phone, the method works equally well.
    Furthermore, for cases when the hand is highly occluded, \meth is able to predict a reasonable pose for it (\eg hand occluded by box and notebook).
    Top row in each example shows the input RGB image, second row shows the predicted posed hands and object projected on the RGB image, and the last row shows the meshes from a different view.
    }
    \label{fig:supp_qual_figure_arctic}
\end{figure*}

In \cref{fig:supp_qual_figure_arctic} we show qualitative results on the ARCTIC dataset~\cite{fan2023arctic}.
\meth estimates pose of both the hands and object in a single forward pass.
Furthermore, \meth works in occlusion, manipulation, and for articulated objects.
Especially in cases where only a portion of the hand is visible (\eg (2, 3) for notebook and (3, 1) for the box) \meth estimates a plausible hand pose for the hand.
Another interesting case to note is of objects like, a phone and a pair of scissors (in location (1, 5), (2, 2), (3, 3), and (3, 4) in \cref{fig:supp_qual_figure_arctic}) where \meth estimates correct pose in minimal object visibility and high occlusion.

\subsection{Scalability of \meth}

In~\cref{fig:num_classes_ablation}, we present plots that show scalability of \meth as the number of classes increase on the ARCTIC dataset.
We here provide all the metrics and the exact numbers from that plot in \cref{tab:samples_merged}.
We report the results tested on an increased test set as classes are added as well as testing on a fixed test set of only 3 classes as more training classes are incorporated. 

\meth is able to generalise from various object classes and improve prediction as the number of classes increase, benefiting from the added diversity despite the increased challenge in predicting poses for more object classes.

\begin{table*}
    \centering
    \caption{\textbf{Effect of training set size under two evaluation protocols}.
    We vary the number of object classes and report results under
    two settings: testing on the same-sized subset, and testing on three object classes.
    Overall, increasing the amount of training data improves most metrics, showing the
    generalisability and learning capability of \meth.
    Note this experiment is done on the ARCTIC dataset.}
    \begin{adjustbox}{width=\textwidth}
    \begin{tabular}{@{}llccccccc@{}}
        \toprule
        \multirow{2}{*}{Protocol} & \multirow{2}{*}{Num. Classes} 
        & \multicolumn{2}{c}{Contact and Relative Positions} 
        & \multicolumn{2}{c}{Motion} 
        & Hand 
        & \multicolumn{2}{c}{Object} \\
        \cmidrule(lr){3-4}\cmidrule(lr){5-6}\cmidrule(lr){7-7}\cmidrule(l){8-9}
        & 
        & CDev [$mm$] $\downarrow$ 
        & MRRPE$_{rl/ro}$ [$mm$] $\downarrow$ 
        & MDev [$mm$] $\downarrow$ 
        & ACC$_{h/o}$ [$m/s^2$] $\downarrow$ 
        & MPJPE [$mm$] $\downarrow$ 
        & AAE [$^{\circ}$] $\downarrow$ 
        & Success Rate [\%] $\uparrow$ \\
        \midrule

        \multirow{4}{*}{\shortstack[l]{Test on \\Num. classes}}
        & $3$ classes         & $51.6$          & $41.9$ / $41.4$          & $18.2$          & $10.3$ / $12.8$          & $27.2$          & $8.2$           & $73.1$ \\
        & $6$ classes         & $46.4$          & $37.4$ / $37.1$          & $14.4$          & $8.4$ / $10.8$           & $23.5$          & $5.5$           & $82.1$ \\
        & $9$ classes         & $33.6$          & $\mathbf{30.0}$ / $30.1$ & $8.2$           & $5.1$ / $6.4$            & $17.7$          & $5.2$           & $\mathbf{87.4}$ \\
        & \meth~($11$ classes)& $\mathbf{31.9}$ & $31.1$ / $\mathbf{29.4}$ & $\mathbf{7.3}$  & $\mathbf{4.8}$ / $\mathbf{6.2}$ & $\mathbf{16.1}$ & $\mathbf{5.0}$  & $82.4$ \\
        \midrule

        \multirow{4}{*}{\shortstack[l]{Test on 3\\object classes}}
        & $3$ classes         & $51.6$          & $41.9$ / $41.4$          & $18.2$          & $10.3$ / $12.8$          & $27.2$          & $8.2$           & $73.1$ \\
        & $6$ classes         & $46.6$          & $36.9$ / $37.6$          & $14.7$          & $8.4$ / $11.3$           & $23.3$          & $7.4$           & $78.6$ \\
        & $9$ classes         & $37.4$          & $\mathbf{31.7}$ / $31.8$ & $8.5$           & $5.1$ / $7.5$            & $18.2$          & $8.0$           & $85.1$ \\
        & \meth~($11$ classes)& $\mathbf{34.5}$ & $32.8$ / $\mathbf{30.6}$ & $\mathbf{7.75}$ & $\mathbf{4.8}$ / $\mathbf{6.7}$ & $\mathbf{16.1}$ & $\mathbf{6.5}$  & $\mathbf{86.1}$ \\
        \bottomrule
    \end{tabular}
    \end{adjustbox}
    \label{tab:samples_merged}
\end{table*}

\subsection{Using predicted classes on \dataset}

In~\cref{tab:ego_epic_results} we show that \meth achieves a classification accuracy of $52.9\%$.
Our main results are reported with oracle class knowledge.
In \cref{tab:auto_results} (subset) we show results on the test samples where the object classifier correctly classifies the object.
\meth achieves similar scores on this subset, using the fully automatic pipeline.
Critically these results are not directly comparable as they are reported on the subset of correctly classified test images.

It is important to highlight that for incorrect classification, evaluation metrics cannot be computed as objects differ in their topology and calculating the object pose for another CAD model is undefined - this requires a vertex level shape matching which is unattainable.

\begin{table}[t]
    \centering
    \caption{\textbf{Fully Automatic Results on \dataset.}
    Results for correct predictions from the object classification layer (subset).
    Compared to results on complete dataset, when we evaluate on the correctly classified subset, \meth performs comparably.}
    \begin{adjustbox}{width=\textwidth}
    \begin{tabular}{@{}lccccccc@{}}
        \toprule
        \multirow{2}{*}{Method} & \multicolumn{2}{c}{Contact and Relative Positions} & \multicolumn{2}{c}{Motion} & Hand & \multicolumn{2}{c}{Object} \\
        \cmidrule(lr){2-3}\cmidrule(lr){4-5}\cmidrule(lr){6-6}\cmidrule(l){7-8}
        & CDev [$mm$] ↓ & MRRPE$_{ro}$ [$mm$] ↓ & MDev [$mm$] ↓ & ACC$_{h/o}$ [$m/s^2$] ↓ & MPJPE [$mm$] ↓ & SR@0.05 [\%] ↑ & SR@0.10 [\%] ↑ \\
        \midrule
        \meth (complete data) & $20.7$ & $65.8$ & $11.4$ & $2.5$/$4.1$ & $19.9$ & $29.8$ & $69.7$  \\
        \meth (subset) & $20.2$ & $64.5$ & $11.4$ & $2.2$/$3.6$ & $18.4$ & $30.4$ & $68.4$  \\
        \bottomrule
    \end{tabular}
    \end{adjustbox}
    \label{tab:auto_results}
\end{table}

\begin{figure}[t]
    \centering
    \includegraphics[width=0.7\linewidth]{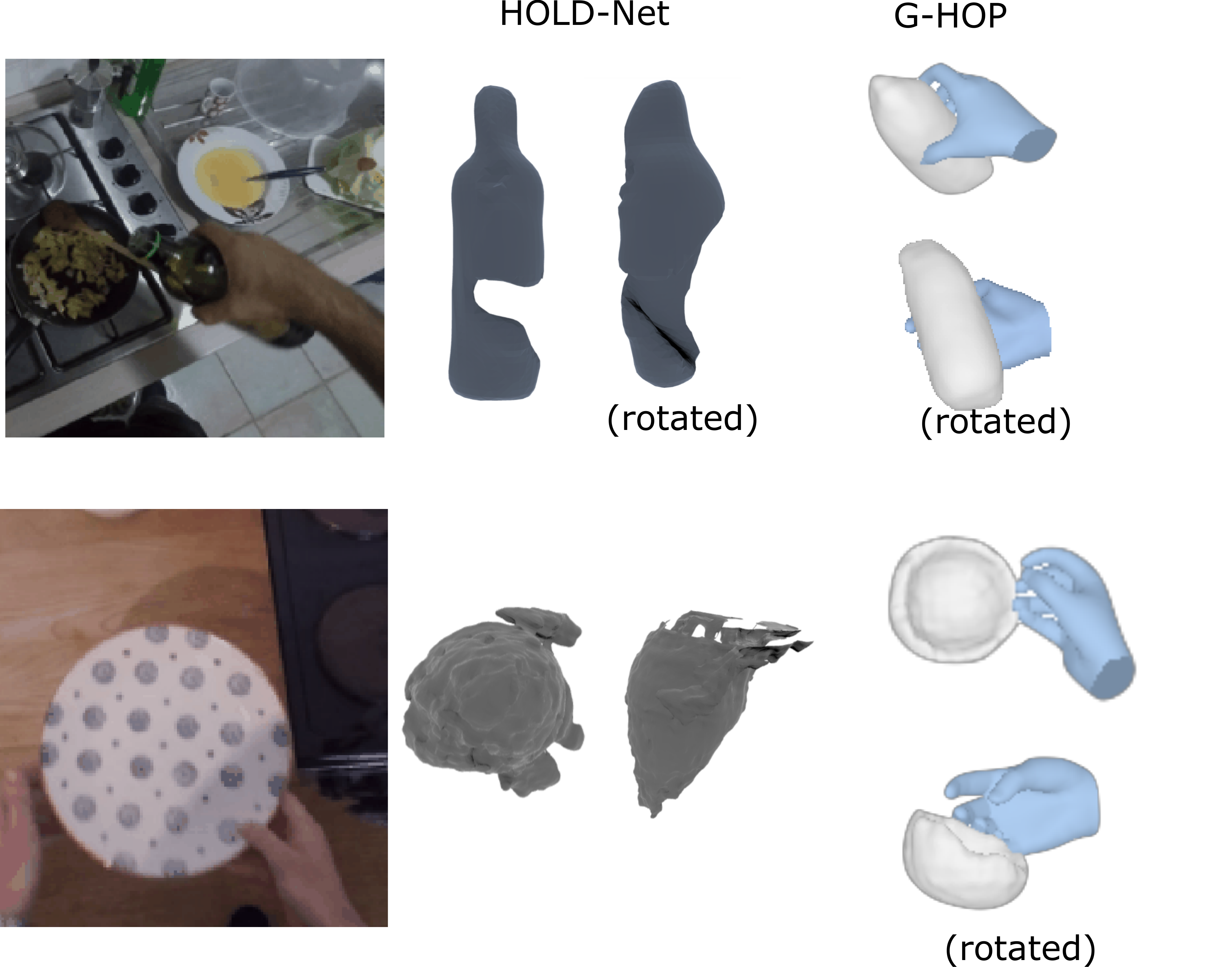}
    \caption{Qualitative examples of CAD-free methods: HOLD-Net~\cite{fan2024hold} and G-HOP~\cite{ye2023ghop}.
    Note that these are video-based methods and hence not directly comparable with \meth.}
    \label{fig:two_cad_free_methods}
\end{figure}

\subsection{Comparing \meth to CAD-Free Methods}

In \cref{fig:two_cad_free_methods}, we explore CAD-free methods and show qualitative results.
We show results from two representative methods: 
i) HOLD-Net~\cite{fan2024hold}, a shape reconstruction method using photo-geometric cues without learnt priors,
ii) G-HOP~\cite{ye2023ghop}, a shape reconstruction method with learnt shape priors.
As shown in~\Cref{fig:two_cad_free_methods}, the shapes or hand-object interactions generated by HOLD-Net and G-HOP deviate from the underlying reality.
Note that the task setting of CAD-free methods is very different from that of~\meth, 
they do not estimate the pose of the object relative to its canonical CAD/shape.

\begin{figure}[t]
    \centering
    \includegraphics[width=\linewidth]{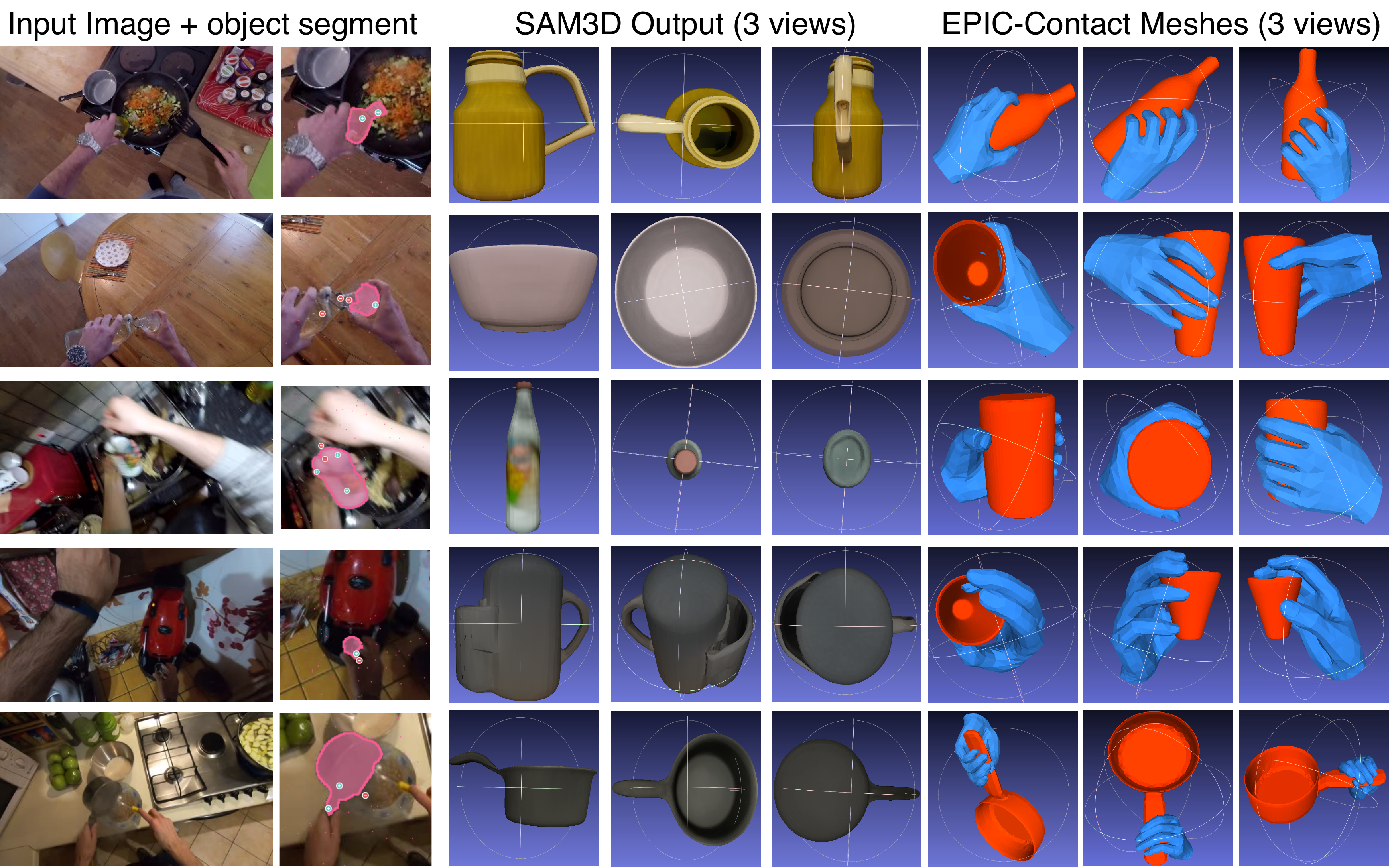}
    \caption{\textbf{SAM3D~\cite{sam3dteam2025sam3d3dfyimages} Failure Cases.}
    SAM3D fails under heavy occlusion and transparent objects.
    Whereas the proposed pipeline for curating the \dataset dataset, despite the challenges, not only provides an accurate object pose, but also a hand pose.
    Notable examples are in rows one, three, and four where SAM3D generates a small container with handle instead of bottle, a bottle instead of can, and a black handled container instead of an espresso cup.
    }
    \label{fig:sam3d_failure}
\end{figure}

Also, we evaluate the recent SAM 3D~\cite{sam3dteam2025sam3d3dfyimages} model for generating 3D CAD models for objects in \dataset.
As shown in \cref{fig:sam3d_failure}, despite training on large amount of data and stronger backbones, SAM 3D fails to generate accurate CAD models for objects in \dataset due to challenges like heavy occlusion and transparent objects.
For example, in the first example, the bottle is predicted as a jug, while in the second example a transparent glass is estimated to be a bowl.
At times, objects are estimated as implausible shapes like the coffee cup in the fourth row.
\meth which assumes a pool of known CAD models enables us to predict accurate posed hand-object meshes.

Using current technology, we argue that our assumption to use fixed object meshes allows exploring in-the-wild pose estimation, whereas advanced backbones and CAD-free approaches still fall short.
In the future, when CAD can be estimated, this can be easily integrated into our approach by predicting the mesh rather than retrieving it.
We believe this exploration requires significant efforts before it's attainable for diverse real scenes.

\subsection{\meth on Exocentric split in ARCTIC}

\begin{table*}[b]
    \centering
    \caption{\textbf{ARCTIC Exocentric Split.}
    Similar to observation on egocentric split, \meth outperforms baselines on majority metrics.
    Especially, CDev and MPJPE show the largest reductions, by $4.6$ and $5.1$, respectively.
    This shows that \meth can generalise to exocentric view as well.
    \textbf{Bold} numbers denote the best performance and numbers in () show the difference of \meth's performance from state-of-the-art.
    }
    \begin{adjustbox}{width=\textwidth}
    \begin{tabular}{@{}lcccccccc@{}}
        \toprule
        \multirow{2}{*}{Method} & \multicolumn{2}{c}{Contact and Relative Positions} & \multicolumn{2}{c}{Motion} & Hand & \multicolumn{3}{c}{Object} \\
        \cmidrule(lr){2-3}\cmidrule(lr){4-5}\cmidrule(lr){6-6}\cmidrule(l){7-9}
        & CDev [$mm$] ↓ & MRRPE$_{rl/ro}$ [$mm$] ↓ & MDev [$mm$] ↓ & ACC$_{h/o}$ [$m/s^2$] ↓ & MPJPE [$mm$] ↓ & AAE [$^{\circ}$] ↓ & SR@0.05 [\%] ↑ & Cls [\%] ↑ \\
        \midrule
        ArcticNet-SF~\cite{fan2023arctic} & $41.4$ & $50.1$ / $37.6$ & $10.4$ & $6.6$ / $8.8$ & $23.0$ & $5.9$ & $71.8$ & - \\
        JointTransformer~\cite{AbouZeid2023JointTransformer} & $29.1$ & $32.9$ / $27.0$ & $8.0$ & $6.0$ / $6.8$ & $17.8$ & $\mathbf{4.1}$ & $86.8$ & - \\
        \meth (\textit{ours}) & $\mathbf{24.5}$ ($-4.6$) & $\mathbf{29.2}$ ($-3.7$) /$\mathbf{23.6}$ ($-3.4$) & $\mathbf{5.4}$ ($-2.6$) & $\mathbf{4.9}$ ($-1.1$) /$\mathbf{4.3}$ ($-2.5$) & $\mathbf{12.7}$ ($-5.1$) & $4.8$ ($+0.7$) & $\mathbf{90.3}$ ($+3.5$) & $\mathbf{99.7}$ \\
        \bottomrule
    \end{tabular}
    \end{adjustbox}
    \label{tab:exo_arctic_results}
\end{table*}

In~\cref{tab:ego_arctic_results}, we report \meth results on the egocentric split of ARCTIC dataset.
Here, we also present results on the exocentric test split.
Consistent with the observation for the egocentric split, the results for exocentric splits improve across various metrics.
As shown in \cref{tab:exo_arctic_results}, metrics for objects, like MDev and MRRPE$_{rl}$ improve by $2.6$ and $3.7$ mm, respectively.
Furthermore, hand reconstruction (MPJPE) improves by $5.1$ mm.
This highlights the generalisation capacity of \meth as well as shows the utility of proposed model to learn from hand priors.

\section{\dataset dataset}
\label{sec:our_dataset}

In this section, we provide additional information on the proposed \dataset dataset.

\subsection{Video Selection}

The videos in \dataset originate from the EPIC-Grasps dataset~\cite{zhu2024grip} as there are challenging and diverse hand-object interactions.
Furthermore, the dataset is paired with 3D meshes for $9$ object categories (mug, pan, glass, cup, saucepan, bottle, plate, bowl, and can) making it ideal for getting posed hand-object meshes.
Additionally, EPIC-Grasps dataset has videos that have ``stable grasp'' between the hand and the object, \ie, when the subject is using an object, the same set of object and hands vertices are in contact.
This allows us to label only one frame per video and then extend that annotation to the rest of the frames using hand pose obtained from WiLoR~\cite{Potamias_2025_CVPR_wilor}.

For hand and object masks, we obtain the ground-truth masks from the VISOR dataset~\cite{darkhalil2022epic}.

\subsection{VLM Scale Estimates and Scale Verification}
The object meshes obtained for the nine objects from~\cite{zhu2024grip} have a fixed scale.
This scale might not match exactly the object instance in the video.
If we use these meshes, without the correct scaling, in our annotation pipeline, the bijective contact obtained would be inaccurate.
To overcome this issue, we infer the scale of the object using a VLM (Gemini 2.5~\cite{comanici2025gemini25pushingfrontier}).
We take the centre frame from the clip and prompt Gemini to provide the scale of the object.

\begin{figure*}[t]
    \centering
    \includegraphics[width=0.9\linewidth]{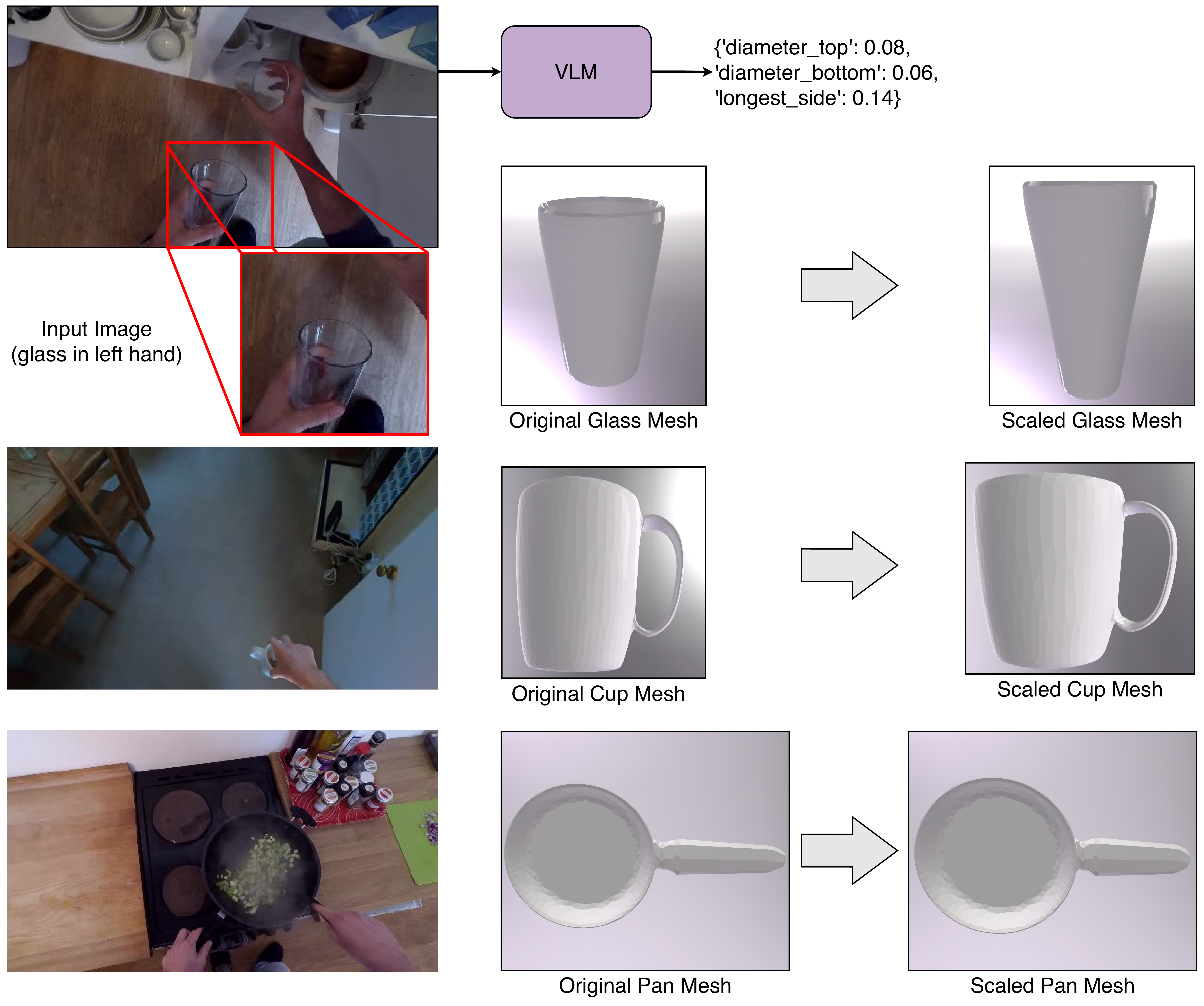}
    \caption{\textbf{Updating Object's Scale}.
    Here we show how the object's scale is updated using VLM's~\cite{comanici2025gemini25pushingfrontier} output to match object's scale in the input image.
    We elaborate the process in the first row where we show the input image, VLM's output, and updated object's (glass in this case) mesh.
    Notice, how the height of the glass changes along with the diameter of base and top to match the glass in input image.
    We show two more examples (excluding VLM and object's zoom for brevity) where height of the cup and diameter of the pan along with its handle length changes.
    }
    \label{fig:objectscale}
\end{figure*}

To get an accurate 3D object mesh, we use different degrees-of-scale enabling non-uniform scaling of the objects.
For example, for a pan, we prompt Gemini to provide both the diameter of the pan (excluding the handle) as well as length of the pan (including the handle).
This allows us to scale the same mesh for pan in two distinct dimensions, allowing us to map to different sizes of pans and varying handle lengths.
\Cref{fig:objectscale} demonstrates how the meshes change upon updating them using the scale provided by the VLM.

Finally, to ensure each object is scaled accordingly, we curate a unique prompt for each object. 
\Cref{section:scale_prompt} contains the prompts used.

\begin{figure}[t]
    \centering
    \includegraphics[width=\linewidth]{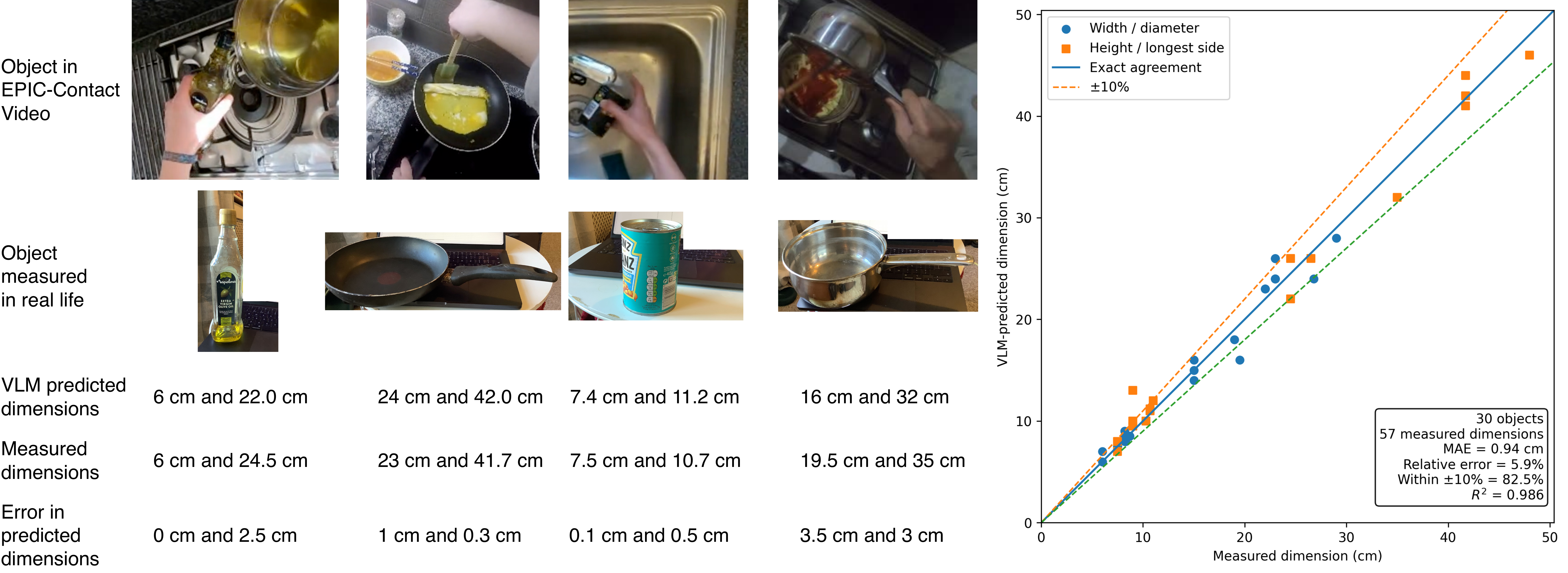}
    \caption{\textbf{VLM scale estimation verification.} We show $4$ representative examples from $30$ objects covering all $9$ classes used to verify these scale predictions. For each example, we show the object in the EPIC-Contact video, the real-life object measured (or an exact size match as in the can example), the VLM predicted dimensions, the measured dimensions, and the error in predicted dimensions. This allows us to evaluate the VLM scale estimates against ground truth dimensions.
    In real measurements, we keep a keyboard in the background for scaling.
    We also show measured vs. VLM-predicted dimensions for all measured dimensions in the 30-object verification set. The solid line denotes exact agreement and the dashed lines denote ±10\% error.
    }
    \label{fig:scale_verification}
\end{figure}

To verify these scale predictions from VLM, we sample $30$ objects covering all $9$ classes and manually compare these to ground truth object sizes.
When possible, we identified objects of a known brand (\eg a specific bottle of oil as shown in \cref{fig:scale_verification}) and measured the dimensions of the same physical object.
For objects with multiple degrees-of-scale, we compare the same dimensions returned by the VLM.
\Cref{fig:scale_verification} shows representative examples from this verification analysis, including the object in the EPIC-Contact video, the same physical object measured in real life, the VLM predicted dimensions, the measured dimensions, and the error in predicted dimensions.
This allows us to evaluate the VLM scale estimates against ground truth dimensions, achieving $0.94$ cm MAE ($5.9\%$ relative error) with $82.5\%$ of samples falling within ±$10\%$ of the true dimensions.

\begin{figure*}
    \centering
    \includegraphics[width=1.0\linewidth]{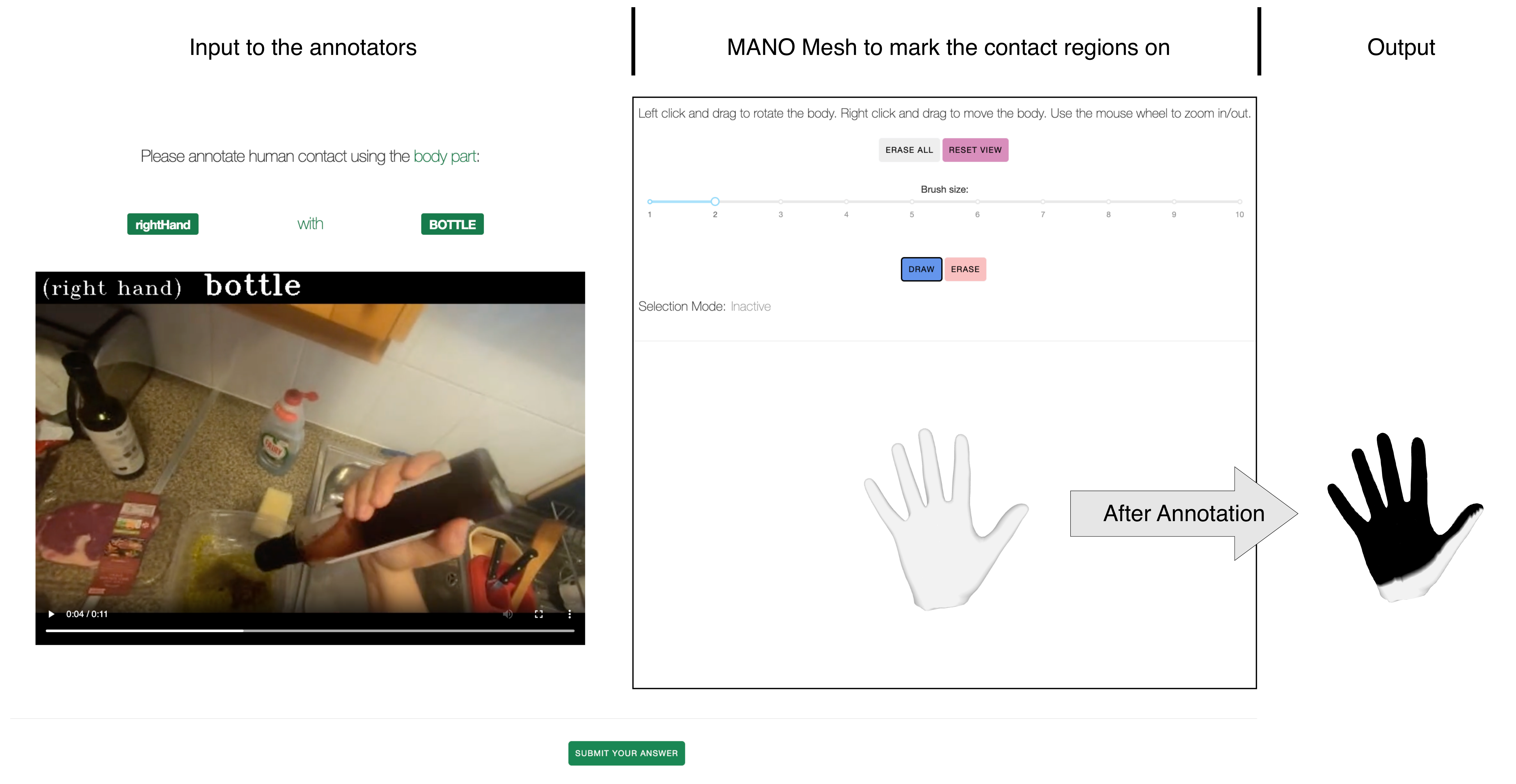}
    \caption{\textbf{Interface to get Hand Contact Regions.}
    The interface is divided into two parts, in the left we show the video to the annotator along with the hand side and object to focus on.
    On the right, we show the MANO mesh along with various controls like zoom, pan, rotate and the paint brush with variable brush size to paint on the mesh.
    For this example, the annotator paints the region on the right hand where the bottle is touching the hand.
    The output is shown in the rightmost column.
    }
    \label{fig:stageoneinterface}
\end{figure*}

\subsection{Annotating Hand Contact Regions}

\begin{figure}
    \centering
    \includegraphics[width=\linewidth]{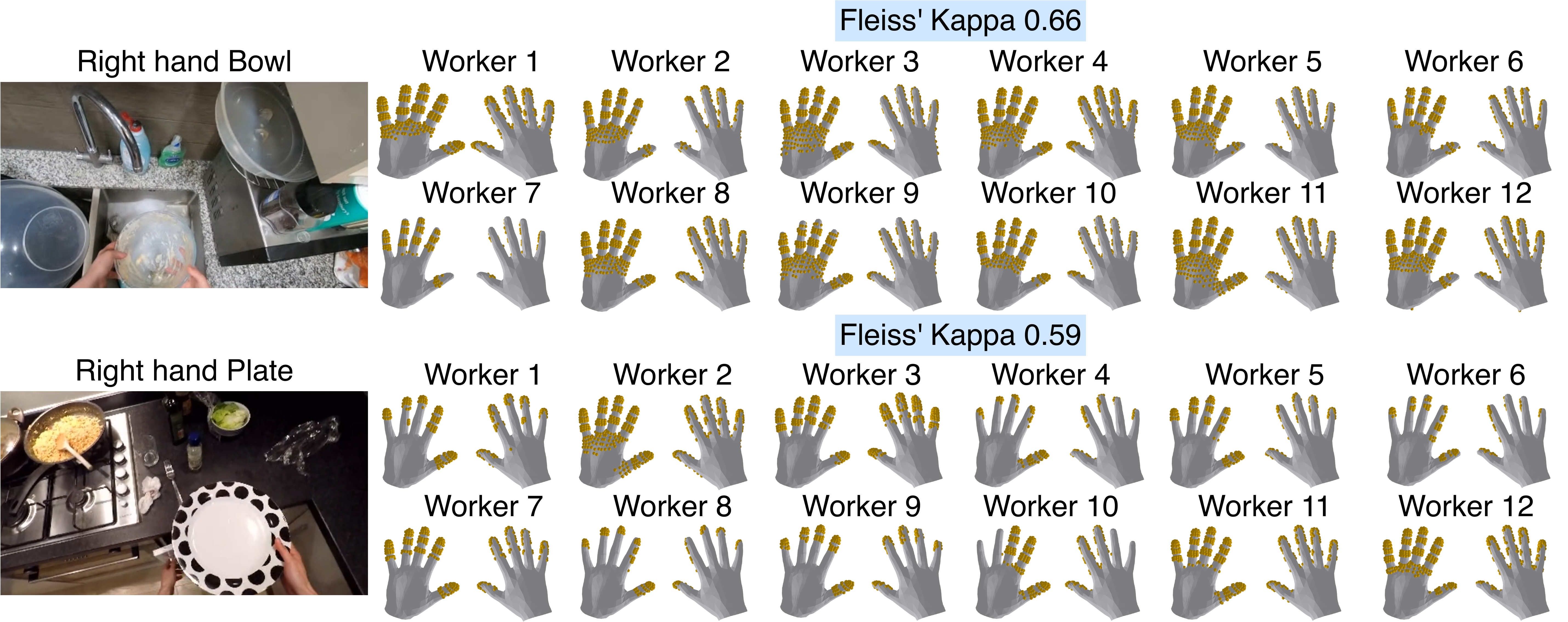}
    \caption{\textbf{Annotator agreement indicated by $\kappa_h$.}
    The figure shows the vertices annotated on the hand (the same hand is shown from the front and the back). We show annotations for all 12 workers. %
    For the bowl example, we get a $\kappa_h$ score of $0.66$ across $12$ workers.
    Most of the annotators agree to the general portion of the hand.
    For the plate example, we get $\kappa_h$ of $0.59$.
    }
    \label{fig:stage_1_kappa}
\end{figure}

\Cref{fig:stageoneinterface} shows the interface created for acquiring the hand contact region.
On the left, we show the video containing a short clip of a stable grasp of one object along with the hand side and object category to focus on (\eg right and bottle in the example given).

To the right of the interface, we show the upscaled MANO mesh~\cite{MANO:SIGGRAPHASIA:2017} for annotators to ``paint'' on.
There are various controls which can be broadly divided into two types, mesh manipulation and painting.
For mesh manipulation, we have the capability to drag, rotate, move, and zoom the hand mesh.
Additionally, there are two buttons, ``ERASE ALL'' to remove all the annotations and ``RESET VIEW'' to bring the mesh to its original position.
For painting, we have various brush sizes to provide better control when painting the contact region on the hand.
We also have two modes ``DRAW'' and ``ERASE'', annotators can select any of these modes, click on the mesh and just hover the cursor over the regions they want to draw/erase.
Allowing drawing/erasing on the mesh with just one click helps with the speed of annotations while maintaining the quality.
Finally, to the right of \cref{fig:stageoneinterface} we show the painted MANO mesh (in black) for this example.

\mypara{Annotations Verification}
We compute the inter-annotator agreement ($\kappa_h$) as done in DECO~\cite{tripathi2023deco} of $0.61$ on $10$ samples annotated by $12$ annotators (as compared to $0.65$ in~\cite{tripathi2023deco}).
As shown in \cref{fig:stage_1_kappa}, the highest $\kappa_h$ of $0.66$ is achieved by the bowl sample and lowest $\kappa_h$ of $0.59$ by the plate sample.

\subsection{Annotating Contact Regions on the Objects}

\begin{figure*}
    \centering
    \includegraphics[width=\linewidth]{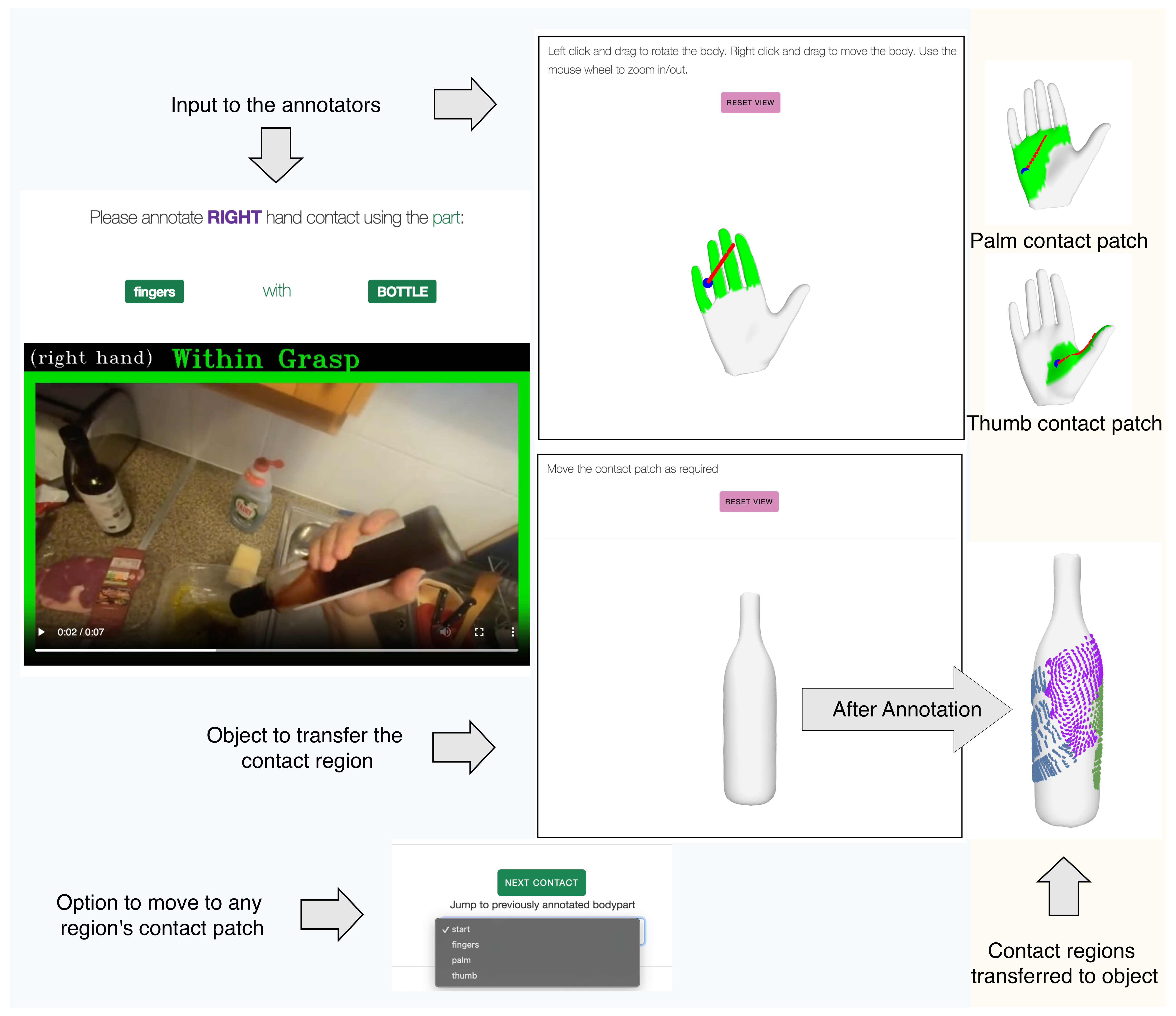}
    \caption{\textbf{Interface to transfer Contact Regions from Hand to Object.}
    Region with blue background shows the interface that annotators see for transferring the hand contact regions to object.
    On the left, we show the video containing the hand-object interaction.
    On the right (top), we show the hand contact region (in green) obtained from the previous exercise along with the calculated ``contact axis'' (blue ball and red line).
    We show the other two contact patches (Palm and Thumb) to the right.
    On the right (bottom), the annotator would transfer the contact axis and eventually the hand contact patch to the object (bottle in this case).
    The interface also contains the option to jump between various contact patches to get more accurate annotations.
    The output is shown in the bottom right of the figure.
    }
    \label{fig:stagetwointerface}
\end{figure*}

\Cref{fig:stagetwointerface} shows a web-based interface to transfer annotated hand regions to the object.
The figure is divided into two parts, the part with blue background shows the interface visible to the annotators, while the yellow background shows additional elements of the interface along with the output.
The interface consists of video that has hand-object interaction, the hand side, and the object to focus on.

On the top-right the interface shows the annotated hand patch (in green) along with the ``contact axis'' (blue ball and red line).
The annotators can rotate, pan, or zoom this MANO mesh to better get sense of the orientation of the patch and contact axis.
There are three such regions (as described in the main paper), fingers, palm, and thumb.
The interface shows one region at a time, but in the figure, we show all three for completion (palm and thumb to the right in yellow). 

At the bottom right in the interface, the annotators see the object mesh on which the hand patch is to be transferred (bottle in \cref{fig:stagetwointerface}).
The annotators make two clicks per patch, first click is to place the blue ball which is the start of the axis and second click to provide direction aligned with the red line.
Once these two clicks are done, the corresponding contact patch is transferred to the object and we obtain bijective mapping of contact points between hand and object.
Similar to the hand, the annotators can rotate, pan, or zoom the object mesh to better position the hand contact patch.
\Cref{fig:stagetwointerface} shows the contact regions transferred on the bottle.
This is also what the annotators see for verification before finalising the annotation.
On average, the annotators take approximately $3$-$4$ minutes per video including the quality verification time.

\begin{figure*}
    \centering
    \includegraphics[width=1.0\linewidth]{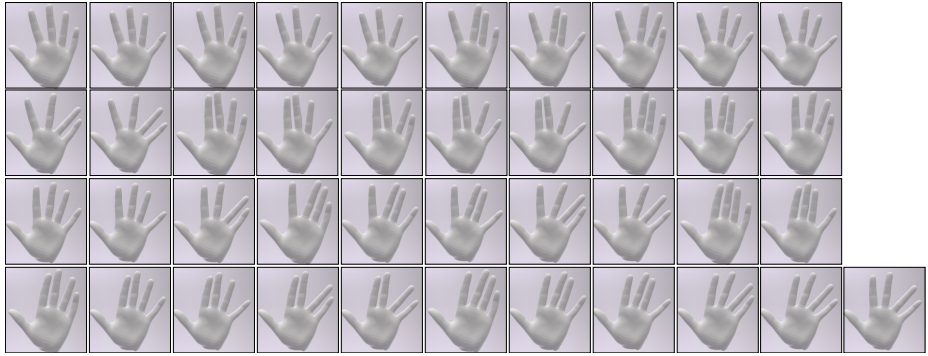}
    \caption{\textbf{$\mathbf{41}$ Flat Hand Poses' Pool}. 
    Set of flat hand poses (left hand in this case) to enable realistic finger distancing when transferring contact patches for four fingers using one contact axis.
    }
    \label{fig:flat_hands}
\end{figure*}

\mypara{Hand finger distance}
Grouping the four fingers into a single region for object transfer poses unique challenges.
We make this decision to ensure consistent transfer.
However, if we use the default MANO flat hand (shown in \cref{fig:stageoneinterface}) to transfer the contact patches, then the distance between the fingers cannot be adjusted.
We overcome this challenge in a novel way while keeping our annotation process efficient.

We identify $41$ common configurations of distances between fingers from the hand pose estimations across the dataset, shown in \cref{fig:flat_hands} for left hand.
These poses represent various finger combinations (\eg. spread, together).
We automatically select the configuration that best matches the hand pose estimation from WiLoR~\cite{Potamias_2025_CVPR_wilor}. We obtain the MANO hand pose vector ($\theta$) and use it to retrieve the closest configuration in the pose vector using geodesic distance.
The flat hand pose with the minimum geodesic distance is used to transfer the contact patch to the object ensuring realistic finger distances are maintained.
To calculate this distance we only use the parts of $\theta \in \mathbb{R}^{48}$ which influence the spread of the fingers, and only calculate the geodesic distance for those dimensions, allowing us to capture the flat hand mesh precisely.

\begin{figure}
    \centering
    \includegraphics[width=\linewidth]{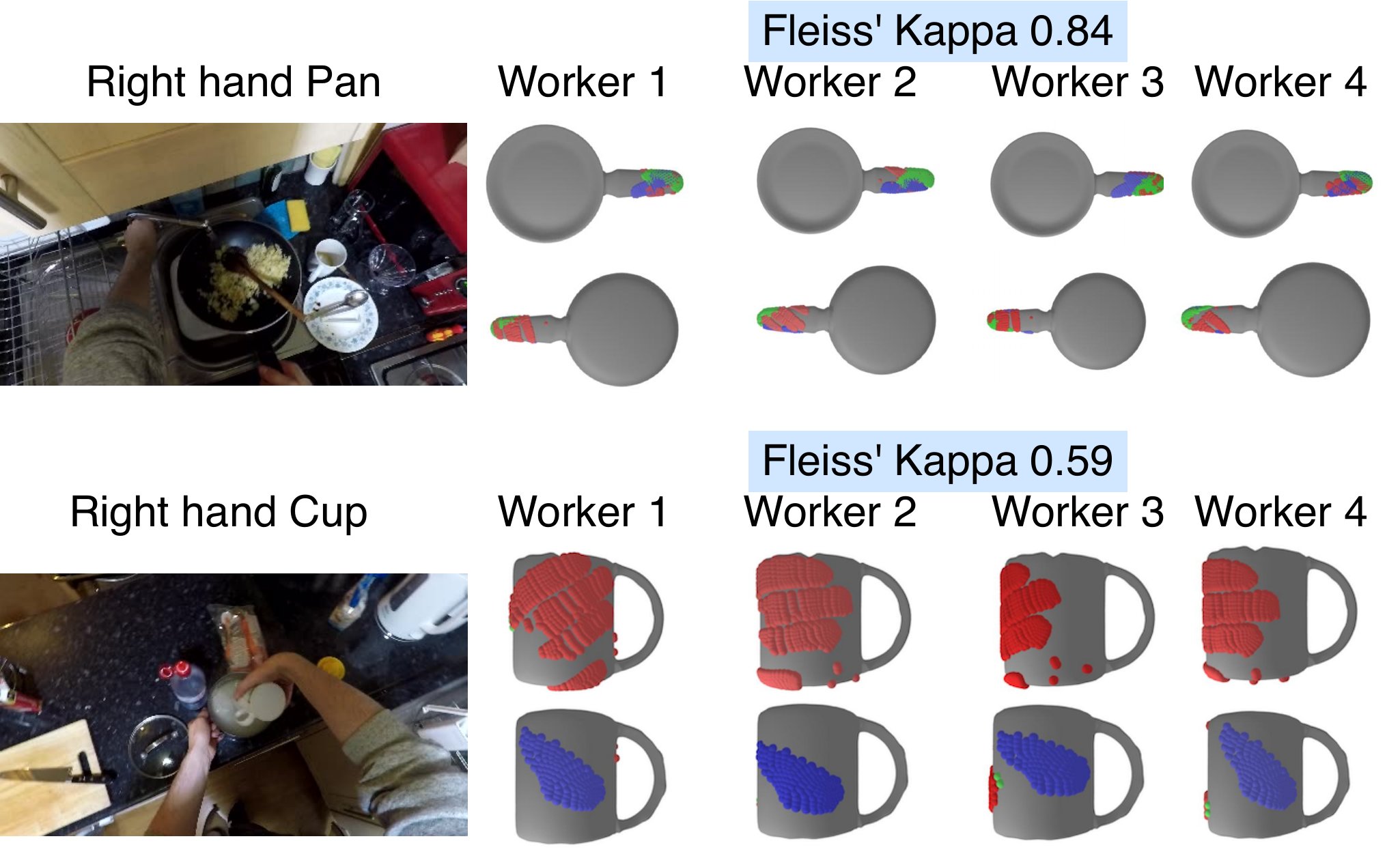}
    \caption{\textbf{Annotator agreement indicated by $\kappa_o$.}
    For the pan example, all annotators mark almost similar regions for fingers (in red), palm (in green), and thumb (in blue).}
    \label{fig:stage_2_kappa}
\end{figure}

\mypara{Annotations Verification}
We compute the inter-annotator agreement ($\kappa_o$) as $0.62$ on $10$ samples annotated by $4$ annotators.
As shown in \cref{fig:stage_2_kappa}, the highest $\kappa_o$ of $0.84$ is achieved by the pan sample and a lower $\kappa_o$ of $0.59$ by the cup sample.
Note that $\kappa$ for bijective correspondences on object is not reported in~\cite{tripathi2023deco,cseke_tripathi_2025_pico}.

\subsection{Details and Evaluation of EC-fit pipeline}

\mypara{Penetration Loss} A penetration loss $\mathcal{L}_p$ is utilised to prevent hand-object penetration. At the stage of refining object pose, we build a Signed Distance Field (SDF) $\Psi_{\mathcal{H}}(\cdot)$ of the hand mesh $\mathcal{H}$, with positive value for a 3D point inside $\mathcal{H}$ and negative value outside $\mathcal{H}$.
Since we aim to penalise object points for being inside the hand mesh, we define the penetration loss as: 
$\mathcal{L}_p^o=\frac{1}{|\mathcal{O}|} \sum_{i=1}^{|\mathcal{O}|} \max(\Psi_{\mathcal{H}}(o_i), 0)$, where $o_i \in \mathcal{O}$ are the object vertices. At the stage of refining hand pose, we instead build the SDF of the object mesh $\mathcal{O}$, and penalise the hand points for being inside the object mesh: $\mathcal{L}_p^h=\frac{1}{|\mathcal{H}|} \sum_{i=1}^{|\mathcal{H}|} 
\max(\Psi_{\mathcal{O}
}(h_i), 0)$, where $h_i \in \mathcal{H}$ are hand vertices.

\mypara{Quality of posed hand-object meshes}
As described in~\cref{section:stagethree}, we utilise the contact regions to obtain posed hand-object meshes for \dataset.
Here, we evaluate the error in this pose estimation, using ARCTIC where ground-truth 3D MoCap poses are available.
We select a random $20\%$ subset of ARCTIC's training set ($37,051$ frames), due to the cost of optimisation and calculate the \textbf{hand contact} vertices and the corresponding \textbf{object contact} vertices.
We then consider this the only annotation present, discard the ground truth 3D object pose, and instead estimate the posed hand and object meshes using our EC-fit pipeline we introduce in~\cref{section:stagethree}.

\Cref{tab:quality_hoi_mesh} shows the estimated error in our posed hand-object meshes when considering our fitting compared to 3D ground truth. The posed meshes achieve a Pose L2 Error (average L2 error of predicted vertices) of $1.9$ mm and a MRRPE of $8.0$ mm, exhibiting close margins to the ground-truth. Additionally, our full pipeline outperforms the results (apart from CDev which increases) over the contact-alignment stage by improving the object pose error.

 \begin{table}[t]
    \centering
    \caption{Quality of Posed Hand-Object Meshes comparing `Contact-based Alignment' (the first part of the pipeline) to the full pipeline.} %
    \begin{adjustbox}{width=0.8\linewidth}
    \begin{tabular}{@{}lccc@{}}
         \toprule
         Method & Pose L2 Error [mm] ↓ & MRRPE [mm] ↓ & CDev [mm] ↓ \\
         \midrule
         EC-fit (contact-alignment only) & $5.0$ & $9.6$ & $\mathbf{7.0}$ \\
         \textbf{EC-fit (full)} & $\mathbf{1.9}$ & $\mathbf{8.0}$ & $12.2$ \\
         \bottomrule
    \end{tabular}
    \end{adjustbox}
    \label{tab:quality_hoi_mesh}
\end{table}

Another way to evaluate our posed meshes is to consider the penetration between the posed hand and object meshes, compared to 3D ground truth poses.
        \Cref{tab:pen_hoi_mesh} compares the MoCap dataset ARCTIC to our posed meshes from EPIC-Contact. While these are different video clips, the average penetration can be considered as an indication of posed meshes quality. The calculation of penetration follows~\cite{jiang2021hand}. The penetration depth and volume of EPIC-Contact is $0.79$ cm and $20.9$ cm$^3$, comparable to ARCTIC meshes which are captured by MoCap devices.
        Evidently, we perform comparably to MoCap in the penetration depth, with increased volume.
        We note that the 3D MoCap also results in considerable penetration volume on average.
        This further showcases the quality of posed hand-object meshes in EPIC-Contact.

 \begin{table}[t]
        \centering
        \captionof{table}{Hand-object penetration values.}
        \resizebox{0.8\linewidth}{!}{
        \begin{tabular}{@{}lcc@{}}
             \toprule
             Dataset & Penetration Volume [$cm^3$] ↓ & Max Penetration Depth [$cm$] ↓ \\
             \midrule
             ARCTIC & \textbf{7.24} & 0.80 \\
             \textbf{EPIC-Contact} & 20.9 & \textbf{0.79} \\
             \bottomrule
        \end{tabular}}
        \label{tab:pen_hoi_mesh}
    \vspace{0pt}
\end{table}

\mypara{Robustness of \meth to posed hand-object meshes}
Next, which is critical for \meth, we evaluate its robustness to the errors in estimated posed meshes.
We run two models, one uses ground truth pose and another uses the estimated posed meshes, from contact regions, using our pipeline. 
We evaluate on the same validation set, with ground truth pose.
The results obtained are in \cref{tab:subset_gt_vs_pico}.
Even with an increase in penetration volume shown in \cref{tab:pen_hoi_mesh}, the metrics remain relatively unchanged, showing that \meth is robust to the small errors in object pose when using contact regions compared to ground truth MoCap.
This further verifies the suitability of the annotations of \dataset to train models for 3D hand-object pose estimation.

\begin{table*}[t]
    \centering
    \caption{\textbf{Robustness to EC-fit pose Noise.}
    On a subset of egocentric split from ARCTIC, we show the effect of using posed hand-object meshes from our EC-fit pipeline.
    \meth when trained using estimated hand-object posed mesh works equally well compared to when training with the ground truth.
    Therefore, \meth is robust to any noise or penetration errors during training.
    }
    \begin{adjustbox}{width=\textwidth}
    \begin{tabular}{@{}lccccccc@{}}
        \toprule
        \multirow{2}{*}{Pose Source} & \multicolumn{2}{c}{Contact and Relative Positions} & \multicolumn{2}{c}{Motion} & Hand & \multicolumn{2}{c}{Object} \\
        \cmidrule(lr){2-3}\cmidrule(lr){4-5}\cmidrule(lr){6-6}\cmidrule(l){7-8}
        & CDev [$mm$] ↓ & MRRPE$_{rl/ro}$ [$mm$] ↓ & MDev [$mm$] ↓ & ACC$_{h/o}$ [$m/s^2$] ↓ & MPJPE [$mm$] ↓ & AAE [$^{\circ}$] ↓ & SR@0.05 [\%] ↑ \\
        \midrule
        Ground Truth  & 40.9 & 35.6 / 37.3 & 11.6 & 6.6 / 10.8 & 20.1 & 7.2 & 71.9 \\
        EC-fit poses & 41.1 & 35.7 / 37.0 & 11.7 & 6.7 / 10.7 & 20.2 & 7.0 & 71.8 \\
        \bottomrule
    \end{tabular}
    \end{adjustbox}
    \label{tab:subset_gt_vs_pico}
\end{table*}

\subsection{Train-test split on \dataset}
We select $10\%$ of the unique videos in \dataset as the test set.
This leaves us with $2{,}035$ videos to train on and $237$ videos to evaluate on out of $2{,}272$ total videos.
We randomly select the $237$ videos while making sure the same videos are not across the train and test split.

\section{Additional Relevant Works}
\label{sec:additional_relevant_works}

In the main paper, we reviewed the most closely related works.
As 3D hand–object interaction understanding is a diverse topic, 
this section discusses additional relevant works in a broader context.

Early work on egocentric hand–object interaction and contact reasoning includes Rogez et al.~\cite{rogez2015understanding}, 
which introduces an egocentric dataset of everyday grasps and focuses on grasp taxonomy and synthetic 3D hand pose generation.
ContactPose~\cite{brahmbhatt2020contactpose} is an in-lab dataset that captures 3D hand–object contact regions using thermal sensors. 
As thermal sensors do not provide point-wise contact correspondences, ContactPose still relies on markers to estimate the object pose.

Another category of relevant work explores hand grasp generation.
The task of hand grasp generation is to produce plausible hand–object interactions, with or without image input as a condition. 
These methods either do not estimate the pose of the CAD model of interest~\cite{ye2023ghop} or do not attempt to reconstruct the interaction observed in the images~\cite{jiang2021hand}.

Regarding datasets, 
we additionally note two early in-lab 3D hand–object datasets:
HO3D~\cite{hampali2020honnotate} and FPHA~\cite{garcia2018first}.
While these two datasets pioneered 3D hand–object annotation,
EPIC-Contact takes a further step in addressing in-the-wild challenges.

\section{Limitations and Future Directions}
\label{sec:limitation_and_future}

For our two proposed contributions: \dataset and \meth, there are some clear limitations, that can be explored in future works.

For \dataset, the annotation pipeline, while robust (as we show in our results), is still time consuming (around $3$-$4$ minutes per annotated centre frame). 
We hypothesise that a better initialisation for contact can be estimated from trained methods to allow annotators to start from a best estimate.
Additionally, we propagate manually-verified ground-truth from a single frame to a clip.
Propagation can be noisy due to errors in hand pose estimates from WiLoR~\cite{Potamias_2025_CVPR_wilor} (hand-side errors or camera placement jumps).
Note that our reported metrics are relative/root-aligned and hence unaffected.
To assist future users, we release per-frame confidence scores by measuring temporal smoothness over the short clip. This allows dataset users to filter confidently labelled propagated frames.

\meth currently estimates poses for a handful of object categories (\eg $9$ for \dataset).
Scaling \meth to more object classes is reserved for future work.
Additionally, while we explore articulated objects in the ARCTIC dataset, we do not explore object articulation in-the-wild which we also leave for future work.

\ifSubfilesClassLoaded{%
  \bibliographystyle{splncs04}
  \bibliography{main}
}{}

\section{Object Scale Prompts}
\label{section:scale_prompt}

\begin{promptbox}
    \textbf{Prompt for Pan} (two degrees-of-scale):\\
    You are an expert AI assistant specialising in visual analysis and real-world object dimension estimation. Analyse the provided image. Your goal is to estimate two key dimensions of the pan being held by the \{left/right\} hand. First dimension is the length of the pan (or the length of the longest side of the pan INCLUDING the handle) and second dimension is the diameter of the pan (EXCLUDING the handle).

    Contextual Information: 1. The human hand is approximately 0.2 metres wide when fully extended. 2. Use the size of the hand as a reference point to estimate the size of the pan. 3. Consider the perspective and any potential occlusions in the image. 4. The camera is worn on the head by the human (this will give you a rough estimate of the distance of the camera from the hand). 5. Because the camera is worn on head it could also make the object look small due to top-down perspective so you might need to compensate for that. 6. Person in the photo is an adult. 7. The pan could be partially occluded by the hand (especially the handle), so you may need to estimate the size based on the visible portion and common knowledge of pan sizes.

    Output Constraints: 1. The output must be a single JSON object. 2. The JSON object must contain two keys: ``diameter'' and ``longest\_side''. 3. The value for each key must be a single floating-point number. 4. The unit for all values must be metres. 5. Example Format: \{``diameter'': 0.28, ``longest\_side'': 0.25\} 6. Do not include any other text, units, or explanations in your final answer.

    Reasoning Process (follow this internally to arrive at your answer): 1. Identify the pan (it is in the hands). 2. Use contextual clues for scale: The primary scale reference is the pair of adult hands holding the pan. 3. Estimate Bounding Box: Mentally fit the tightest possible 3D rectangular box around the pan. Estimate its three dimensions (e.g., length, width, height) based on visual evidence and common knowledge of the object. 4. Determine the longest edge: Identify the largest of the three estimated bounding box dimensions. Make sure this includes the handle of the pan. This value is your ``longest\_side''. 5. Estimate Diameter of the pan's top. Use the knowledge of standard pan sizes to estimate it. Do NOT include the handle in this estimation. 6. Synthesise and Finalise: Combine the visual information with common knowledge about typical pan sizes to produce the most plausible final estimates for both required dimensions. 7. Convert and Format: Ensure both final estimates are in metres and format them into the specified JSON structure.
\end{promptbox}

\begin{promptbox}
    \textbf{Prompt for Bottle} (two degrees-of-scale):\\
    You are an expert AI assistant specialising in visual analysis and real-world object dimension estimation. Analyse the provided image. Your goal is to estimate two key dimensions of the bottle being held by the \{left/right\} hand. First dimension is the height of the bottle (or the length of the longest side of the bottle) and second dimension is the diameter of the base of the bottle.

    Contextual Information: 1. The human hand is approximately 0.2 metres wide when fully extended. 2. Use the size of the hand as a reference point to estimate the size of the object. 3. Consider the perspective and any potential occlusions in the image. 4. The camera is worn on the head by the human (this will give you a rough estimate of the distance of the camera from the hand; this could also make the object look small due to top-down perspective so keep that in mind). 5. Person in the photo is an adult. 6. The bottle could be partially occluded by the hand, so you may need to estimate the size based on the visible portion and common knowledge of bottle sizes. 7. To compensate for the top-down perspective, first estimate the more reliable diameter, then infer the true height by considering that a bottle could be typically 3 to 4 times taller than it is wide.

    Output Constraints: 1. The output must be a single JSON object. 2. The JSON object must contain two keys: ``diameter'' and ``longest\_side''. 3. The value for each key must be a single floating-point number. 4. The unit for all values must be metres. 5. Example Format: {``diameter'': 0.28, ``longest\_side'': 0.25} 6. Do not include any other text, units, or explanations in your final answer.

    Reasoning Process (follow this internally to arrive at your answer): 1. Identify the bottle (it is in the hands). 2. Use contextual clues for scale: The primary scale reference is the pair of adult hands holding the bottle. 3. Estimate Bounding Box: Mentally fit the tightest possible 3D rectangular box around the object. Estimate its three dimensions (e.g., length, width, height) based on visual evidence and common knowledge of the object. 4. Determine Longest Edge: Identify the largest of the three estimated bounding box dimensions. This value is your ``longest\_side''. 5. Estimate Diameter of the bottle. You could use the curling of the fingers around the bottle for cues. 6. Synthesise and Finalise: Combine the visual information with common knowledge about typical bottle sizes to produce the most plausible final estimates for both required dimensions. 7. Convert and Format: Ensure both final estimates are in metres and format them into the specified JSON structure.
\end{promptbox}

\begin{promptbox}
    \textbf{Prompt for Bowl} (two degrees-of-scale):\\
    You are an expert AI assistant specialising in visual analysis and real-world object dimension estimation. Analyse the provided image. Your goal is to estimate two key dimensions of the bowl being held by the \{left/right\} hand. First dimension is the depth of the bowl (imagine there is a lid on the top; what would be the distance from the lid to the bottom of the bowl) and second dimension is the diameter of the top of the bowl (the opening part).

    Contextual Information: 1. The human hand is approximately 0.2 metres wide when fully extended. 2. Use the size of the hand as a reference point to estimate the size of the bowl. 3. Consider the perspective and any potential occlusions in the image. 4. The camera is worn on the head by the human (this will give you a rough estimate of the distance of the camera from the hand; this could also make the object look small due to top-down perspective so keep that in mind). 5. Person in the photo is an adult. 6. The bowl could be partially occluded by the hand, so you may need to estimate the size based on the visible portion and common knowledge of bowl sizes. 7. Because the camera is worn on head it could also make the object look small due to top-down perspective so you might need to compensate for that.

    Output Constraints: 1. The output must be a single JSON object. 2. The JSON object must contain two keys: ``diameter'' and ``longest\_side''. 3. The value for each key must be a single floating-point number. 4. The unit for all values must be metres. 5. Example Format: {``diameter'': 0.28, ``longest\_side'': 0.25} 6. Do not include any other text, units, or explanations in your final answer.

    Reasoning Process (follow this internally to arrive at your answer): 1. Identify the bowl (it is in the hands). 2. Use contextual clues for scale: The primary scale reference is the pair of adult hands holding the bowl. 3. Estimate Bounding Box: Mentally fit the tightest possible 3D rectangular box around the bowl. Estimate its three dimensions (e.g., length, width, height) based on visual evidence and common knowledge of the object. 4. Determine the depth: Identify the depth of the bowl from the estimated bounding box dimensions. This value is your ``longest\_side''. 5. Estimate Diameter of the bowl. You could use the curling of the fingers around the bowl for cues. 6. Synthesise and Finalise: Combine the visual information with common knowledge about typical bowl sizes to produce the most plausible final estimates for both required dimensions. 7. Convert and Format: Ensure both final estimates are in metres and format them into the specified JSON structure.
\end{promptbox}

\begin{promptbox}
    \textbf{Prompt for Can} (two degrees-of-scale):\\
    You are an expert AI assistant specialising in visual analysis and real-world object dimension estimation. Analyse the provided image. Your goal is to estimate two key dimensions of the can being held by the \{left/right\} hand. First dimension is the height of the can (or the length of the longest side of the can) and second dimension is the diameter of the top/base of the can.

    Contextual Information: 1. The human hand is approximately 0.2 metres wide when fully extended. 2. Use the size of the hand as a reference point to estimate the size of the can. 3. Consider the perspective and any potential occlusions in the image. 4. The camera is worn on the head by the human (this will give you a rough estimate of the distance of the camera from the hand; this could also make the object look small due to top-down perspective so keep that in mind). 5. Person in the photo is an adult. 6. The can could be partially occluded by the hand, so you may need to estimate the size based on the visible portion and common knowledge of can sizes. 7. Because the camera is worn on head it could also make the object look small due to top-down perspective so you might need to compensate for that.

    Output Constraints: 1. The output must be a single JSON object. 2. The JSON object must contain two keys: ``diameter'' and ``longest\_side''. 3. The value for each key must be a single floating-point number. 4. The unit for all values must be metres. 5. Example Format: {``diameter'': 0.28, ``longest\_side'': 0.25} 6. Do not include any other text, units, or explanations in your final answer.

    Reasoning Process (follow this internally to arrive at your answer): 1. Identify the can (it is in the hands). 2. Use contextual clues for scale: The primary scale reference is the pair of adult hands holding the can. 3. Estimate Bounding Box: Mentally fit the tightest possible 3D rectangular box around the can. Estimate its three dimensions (e.g., length, width, height) based on visual evidence and common knowledge of the object. 4. Determine the longest edge: Identify the largest of the three estimated bounding box dimensions. This value is your ``longest\_side''. 5. Estimate Diameter of the can. You could use the curling of the fingers around the can for cues. 6. Synthesise and Finalise: Combine the visual information with common knowledge about typical can sizes to produce the most plausible final estimates for both required dimensions. 7. Convert and Format: Ensure both final estimates are in metres and format them into the specified JSON structure.
\end{promptbox}

\begin{promptbox}
    \textbf{Prompt for Cup} (two degrees-of-scale):\\
    You are an expert AI assistant specialising in visual analysis and real-world object dimension estimation. Analyse the provided image. Your goal is to estimate two key dimensions of the cup being held by the \{left/right\} hand. First dimension is the height of the cup (or the length of the longest side of the cup) and second dimension is the diameter of the top of the cup.

    Contextual Information: 1. The human hand is approximately 0.2 metres wide when fully extended. 2. Use the size of the hand as a reference point to estimate the size of the cup. 3. Consider the perspective and any potential occlusions in the image. 4. The camera is worn on the head by the human (this will give you a rough estimate of the distance of the camera from the hand; this could also make the object look small due to top-down perspective so keep that in mind). 5. Person in the photo is an adult. 6. The cup could be partially occluded by the hand, so you may need to estimate the size based on the visible portion and common knowledge of cup sizes. 7. Because the camera is worn on head it could also make the object look small due to top-down perspective so you might need to compensate for that. 8. Ignore the handle of the cup (if present).

    Output Constraints: 1. The output must be a single JSON object. 2. The JSON object must contain two keys: ``diameter'' and ``longest\_side''. 3. The value for each key must be a single floating-point number. 4. The unit for all values must be metres. 5. Example Format: {``diameter'': 0.28, ``longest\_side'': 0.25} 6. Do not include any other text, units, or explanations in your final answer.

    Reasoning Process (follow this internally to arrive at your answer): 1. Identify the cup (it is in the hands). 2. Use contextual clues for scale: The primary scale reference is the pair of adult hands holding the cup. 3. Estimate Bounding Box: Mentally fit the tightest possible 3D rectangular box around the cup. Estimate its three dimensions (e.g., length, width, height) based on visual evidence and common knowledge of the object. 4. Determine the longest edge: Identify the largest of the three estimated bounding box dimensions. This value is your ``longest\_side''. 5. Estimate Diameter of the cup's top. You could use the curling of the fingers around the cup for cues. 6. Synthesise and Finalise: Combine the visual information with common knowledge about typical cup sizes to produce the most plausible final estimates for both required dimensions. 7. Convert and Format: Ensure both final estimates are in metres and format them into the specified JSON structure.
\end{promptbox}

\begin{promptbox}
    \textbf{Prompt for Glass} (three degrees-of-scale):\\
    You are an expert AI assistant specialising in visual analysis and real-world object dimension estimation. Analyse the provided image. Your goal is to estimate three key dimensions of the glass being held by the \{left/right\} hand. First dimension is the height of the glass (or the length of the longest side of the glass), second dimension is the diameter of the top of the glass and third is the diameter of the bottom of the glass.

    Contextual Information: 1. The human hand is approximately 0.2 metres wide when fully extended. 2. Use the size of the hand as a reference point to estimate the size of the glass. 3. Consider the perspective and any potential occlusions in the image. 4. The camera is worn on the head by the human (this will give you a rough estimate of the distance of the camera from the hand; this could also make the object look small due to top-down perspective so keep that in mind). 5. Person in the photo is an adult. 6. The glass could be partially occluded by the hand, so you may need to estimate the size based on the visible portion and common knowledge of glass sizes. 7. Because the camera is worn on head it could also make the object look small due to top-down perspective so you might need to compensate for that.

    Output Constraints: 1. The output must be a single JSON object. 2. The JSON object must contain three keys: ``diameter\_top'', ``diameter\_bottom'', and ``longest\_side''. 3. The value for each key must be a single floating-point number. 4. The unit for all values must be metres. 5. Example Format: {``diameter\_top'': 0.28, ``diameter\_bottom'': 0.26, ``longest\_side'': 0.25} 6. Do not include any other text, units, or explanations in your final answer.

    Reasoning Process (follow this internally to arrive at your answer): 1. Identify the glass (it is in the hands). 2. Use contextual clues for scale: The primary scale reference is the pair of adult hands holding the glass. 3. Estimate Bounding Box: Mentally fit the tightest possible 3D rectangular box around the glass. Estimate its three dimensions (e.g., length, width, height) based on visual evidence and common knowledge of the object. 4. Determine the longest edge: Identify the largest of the three estimated bounding box dimensions. This value is your ``longest\_side''. 5. Estimate Diameter of the glass's top. You could use the curling of the fingers around the glass for cues. 6. Synthesise and Finalise: Combine the visual information with common knowledge about typical glass sizes to produce the most plausible final estimates for both required dimensions. 7. Convert and Format: Ensure both final estimates are in metres and format them into the specified JSON structure.
\end{promptbox}

\begin{promptbox}
    \textbf{Prompt for Mug} (two degrees-of-scale):\\
    You are an expert AI assistant specialising in visual analysis and real-world object dimension estimation. Analyse the provided image. Your goal is to estimate two key dimensions of the mug being held by the \{left/right\} hand. First dimension is the height of the mug (or the length of the longest side of the mug) and second dimension is the diameter of the top of the mug.

    Contextual Information: 1. The human hand is approximately 0.2 metres wide when fully extended. 2. Use the size of the hand as a reference point to estimate the size of the mug. 3. Consider the perspective and any potential occlusions in the image. 4. The camera is worn on the head by the human (this will give you a rough estimate of the distance of the camera from the hand). 5. Because the camera is worn on head it could also make the object look small due to top-down perspective so you might need to compensate for that. 6. Person in the photo is an adult. 7. The mug could be partially occluded by the hand, so you may need to estimate the size based on the visible portion and common knowledge of mug sizes. 8. Ignore the handle of the mug (if present).

    Output Constraints: 1. The output must be a single JSON object. 2. The JSON object must contain two keys: ``diameter'' and ``longest\_side''. 3. The value for each key must be a single floating-point number. 4. The unit for all values must be metres. 5. Example Format: {``diameter'': 0.28, ``longest\_side'': 0.25} 6. Do not include any other text, units, or explanations in your final answer.

    Reasoning Process (follow this internally to arrive at your answer): 1. Identify the mug (it is in the hands). 2. Use contextual clues for scale: The primary scale reference is the pair of adult hands holding the mug. 3. Estimate Bounding Box: Mentally fit the tightest possible 3D rectangular box around the mug. Estimate its three dimensions (e.g., length, width, height) based on visual evidence and common knowledge of the object. 4. Determine the longest edge: Identify the largest of the three estimated bounding box dimensions. This value is your ``longest\_side''. 5. Estimate Diameter of the mug's top. You could use the curling of the fingers around the mug for cues. 6. Synthesise and Finalise: Combine the visual information with common knowledge about typical mug sizes to produce the most plausible final estimates for both required dimensions. 7. Convert and Format: Ensure both final estimates are in metres and format them into the specified JSON structure.
\end{promptbox}

\begin{promptbox}
    \textbf{Prompt for Saucepan} (two degrees-of-scale):\\
    You are an expert AI assistant specialising in visual analysis and real-world object dimension estimation. Analyse the provided image. Your goal is to estimate two key dimensions of the saucepan being held by the \{left/right\} hand. First dimension is the length of the saucepan (or the length of the longest side of the saucepan INCLUDING the handle) and second dimension is the diameter of the saucepan (EXCLUDING the handle).

    Contextual Information: 1. The human hand is approximately 0.2 metres wide when fully extended. 2. Use the size of the hand as a reference point to estimate the size of the saucepan. 3. Consider the perspective and any potential occlusions in the image. 4. The camera is worn on the head by the human (this will give you a rough estimate of the distance of the camera from the hand). 5. Because the camera is worn on head it could also make the object look small due to top-down perspective so you might need to compensate for that. 6. Person in the photo is an adult. 7. The saucepan could be partially occluded by the hand (especially the handle), so you may need to estimate the size based on the visible portion and common knowledge of saucepan sizes.

    Output Constraints: 1. The output must be a single JSON object. 2. The JSON object must contain two keys: ``diameter'' and ``longest\_side''. 3. The value for each key must be a single floating-point number. 4. The unit for all values must be metres. 5. Example Format: {``diameter'': 0.28, ``longest\_side'': 0.25} 6. Do not include any other text, units, or explanations in your final answer.

    Reasoning Process (follow this internally to arrive at your answer): 1. Identify the saucepan (it is in the hands). 2. Use contextual clues for scale: The primary scale reference is the pair of adult hands holding the saucepan. 3. Estimate Bounding Box: Mentally fit the tightest possible 3D rectangular box around the saucepan. Estimate its three dimensions (e.g., length, width, height) based on visual evidence and common knowledge of the object. 4. Determine the longest edge: Identify the largest of the three estimated bounding box dimensions. Make sure this includes the handle of the saucepan. This value is your ``longest\_side''. 5. Estimate Diameter of the saucepan's top. Use the knowledge of standard saucepan sizes to estimate it. Do NOT include the handle in this estimation. 6. Synthesise and Finalise: Combine the visual information with common knowledge about typical saucepan sizes to produce the most plausible final estimates for both required dimensions. 7. Convert and Format: Ensure both final estimates are in metres and format them into the specified JSON structure.
\end{promptbox}

\begin{promptbox}
    \textbf{Prompt for Plate} (one degree-of-scale):\\
    You are an expert AI assistant specialising in visual analysis and real-world object dimension estimation. Analyse the provided image. Your goal is to estimate the diameter of the plate being held by the \{left/right\} hand.

    Contextual Information: 1. The human hand is approximately 0.2 metres wide when fully extended. 2. Use the size of the hand as a reference point to estimate the size of the plate. 3. Consider the perspective and any potential occlusions in the image. 4. The camera is worn on the head by the human (this will give you a rough estimate of the distance of the camera from the hand). 5. Because the camera is worn on head it could also make the object look small due to top-down perspective so you might need to compensate for that. 6. Person in the photo is an adult. 7. The plate could be partially occluded by the hand, so you may need to estimate the diameter based on the visible portion and common knowledge of plate sizes.

    Output Constraints: 1. The output must be a single JSON object. 2. The JSON object must contain one key: ``diameter''. 3. The value for the key must be a single floating-point number. 4. The unit for the value must be metres. 5. Example Format: {``diameter'': 0.28} 6. Do not include any other text, units, or explanations in your final answer.

    Reasoning Process (follow this internally to arrive at your answer): 1. Identify the plate (it is in the hands). 2. Use contextual clues for scale: The primary scale reference is the pair of adult hands holding the plate. 3. Estimate Bounding Box: Mentally fit the tightest possible 3D rectangular box around the plate. Estimate its three dimensions (e.g., length, width, height) based on visual evidence and common knowledge of the object. 4. Estimate the diameter of the plate. You could use the curling of the fingers around the plate for cues. 6. Synthesise and Finalise: Combine the visual information with common knowledge about typical plate sizes to produce the most plausible final estimates for the required dimension. 7. Convert and Format: Ensure both final estimates are in metres and format them into the specified JSON structure.
\end{promptbox}

\end{document}

\end{document}